\documentclass{article}

\usepackage[final, nonatbib]{neurips_2022}

\usepackage[utf8]{inputenc} 
\usepackage[T1]{fontenc}    
\usepackage{hyperref}       
\usepackage{url}            
\usepackage{booktabs}       
\usepackage{amsfonts}       
\usepackage{nicefrac}       
\usepackage{microtype}      
\usepackage[table]{xcolor}         

\usepackage{amsthm}
\usepackage{graphicx}
\usepackage{subcaption}
\usepackage{bbm}
\usepackage{amsmath}

\DeclareMathOperator*{\argmin}{arg\,min}
\usepackage{mathtools}
\usepackage{algorithm}
\usepackage[noend]{algpseudocode}

\usepackage{lipsum} 
\usepackage{booktabs} 

\usepackage[table]{xcolor}
\newcommand{\algnamenospace}{MABSplit}
\newcommand{\algname}{MABSplit }

\newtheorem{theorem}{Theorem}

\newcommand{\aln}[1]{\begin{align*}#1\end{align*}}

\newcounter{numcount}
\newif\iflong
\longfalse

\newif\ifdraft
\drafttrue

\renewcommand{\paragraph}[1]{\noindent {\bf #1}}

\title{\algnamenospace: Faster Forest Training Using Multi-Armed Bandits}

\author{%
  Mo Tiwari$^1$  \\
  \And
  Ryan Kang$^{1,*}$ \\
  \And
  Je-Yong Lee$^{2,*}$  \\
  \And
  Sebastian Thrun$^1$ \\
  \And Chris Piech$^1$ \\
  \And
  Ilan Shomorony$^{\#,3}$\\
  \And
  Martin Jinye Zhang$^{\#,4}$ \\
}

\begin{document}

\maketitle


\begin{abstract}
\label{sec:abstract}

Random forests are some of the most widely used machine learning models today, especially in domains that necessitate interpretability.
We present an algorithm that accelerates the training of random forests and other popular tree-based learning methods.
At the core of our algorithm is a novel node-splitting subroutine, dubbed \algnamenospace, used to efficiently find split points when constructing decision trees.
Our algorithm borrows techniques from the multi-armed bandit literature to judiciously determine how to allocate samples and computational power across candidate split points. 
We provide theoretical guarantees that \algname improves the sample complexity of each node split from linear to logarithmic in the number of data points. 
In some settings, \algname leads to 100x faster training (an 99\% reduction in training time) without any decrease in generalization performance. 
We demonstrate similar speedups when \algname is used across a variety of forest-based variants, such as Extremely Random Forests and Random Patches. 
We also show our algorithm can be used in both classification and regression tasks.
Finally, we show that \algname outperforms existing methods in generalization performance and feature importance calculations under a fixed computational budget. 
All of our experimental results are reproducible via a one-line script at \href{https://github.com/ThrunGroup/FastForest} {\texttt{https://github.com/ThrunGroup/FastForest}}.

\end{abstract}


\section{Introduction \label{sec:intro}}
\let\thefootnote\relax\footnotetext{$*$ denotes equal contribution. $\#$ denotes joint supervision. Correspondence should be addressed to M.T. (\texttt{motiwari@stanford.edu}), I.S. (\texttt{ilans@illinois.edu}), and M.J.Z. (\texttt{jinyezhang@hsph.harvard.edu}).}
\let\thefootnote\relax\footnotetext{1: Department of Computer Science, Stanford University}
\let\thefootnote\relax\footnotetext{2: Oxford University}
\let\thefootnote\relax\footnotetext{3: Electrical and Computer Engineering, University of Illinois at Urbana-Champaign}
\let\thefootnote\relax\footnotetext{4: Department of Epidemiology, Harvard T.H. Chan School of Public Health}



Random Forest (RF) is a supervised learning technique that is widely used for classification and regression tasks \cite{ho1995random,breiman2001random}.
In RF, an ensemble of decision trees (DTs) is trained for the same prediction task.
Each DT consists of a series of nodes that represent \texttt{if/then/else} comparisons on the feature values of a given datapoint and are used to produce an output label.
In RF, each DT is typically trained on a \emph{bootstrap} sample of the original dataset and considers a random sample of available features at each node split \cite{breiman1996bagging}. 
The prediction of the each DT is aggregated to provide an output label for the whole RF.
By aggregating the prediction of each DT in the ensemble, RFs tend to be more robust to noise and overfitting  \cite{rokach2005top} and are capable of capturing more complex patterns in the data than a single DT \cite{tumer1996error}.


RF has gained tremendous popularity due to its flexibility, usefulness in multi-class classification and regression tasks, 
high performance across a broad range of data types, natural support for missing features, and relatively low computational complexity \cite{yates2021fastforest,silveira2019object,benali2019solar,lakshmanaprabu2019random,james2013introduction}.
Furthermore, RF inherits the interpretability of decision trees because the prediction of each constituent DT can be explained through a sequence of binary decisions.
RFs have been successfully applied in contexts as varied as the prediction of legal court decisions \cite{katz2017general},  solar radiation analysis \cite{benali2019solar}, and the Higgs Boson classification problem~\cite{azhari2020higgs}.
In the era of big data, a simple and flexible machine learning technique such as RF is expected to play a key role in processing large datasets and providing accurate and interpretable predictions.


The need for training prediction models on massive datasets and doing so on compute-constrained hardware, such as smartphones and Internet-of-Things devices, requires the development of new algorithms that can deliver faster results without sacrificing generalization performance~\cite{yates2021fastforest}.
For this reason, recent work has proposed ways to accelerate the training of RFs, both at the algorithmic level and at the hardware level.

At the algorithmic level, most work focuses on fast construction of each individual DT. 
Each DT is built by identifying the feature $f$ and threshold $t$ that best split the data points according to the prediction targets. 
The data points are split into subsets based on whether their feature $f$ has a value less than $t$ or feature $f$ has a value greater than $t$. The process is then recursed for each resulting subset.
Most of the complexity in this process is in identifying the pair $(f,t)$ that provides the best split for a set of $N$ data points, which typically requires $O(N)$ computation per split.
Recent proposals include computing (or estimating) $f$ and $t$ from a subsample of the data points and features, or quantizing the feature values. The latter technique creates a histogram of values of each feature across the data points and restricts $t$ to be at the edges of histogram bins.


While existing approaches provide significant speed up in the training of RFs, they often require prespecification of fixed hyperparameters, such as the proportion of data points or features to subsample, and are not adaptive to the underlying data distribution.
Moreover, when comparing different candidate features for a split, all features are treated on equal footing and the quality of their split is computed based on the same number of data points.
Intuitively, this is wasteful because features that are not informative for the prediction task can be identified based on a smaller number of data points. 
Alternatively, an adaptive scheme could better allocate computational resources towards a promising set of candidate features and achieve a better tradeoff between computational cost and generalization performance.

In this work, we propose \algnamenospace, a fast subroutine for the node-splitting problem, which adaptively refines the estimate of the ``quality'' of each feature-threshold pair $(f,t)$ as a candidate split. 
Bad split candidates can be discarded early, which can lead to significant computational savings.
The core idea behind our algorithm is to formulate the node-splitting task as a multi-armed bandit problem \cite{lai1985asymptotically,audibert2010best,bagaria2018medoids,tiwari2020banditpam}, where each pair $(f,t)$ is a distinct arm.
The unknown parameter of each arm, $\mu_{ft}$, corresponds to the quality of the split based on feature $f$ and threshold $t$, where the split quality is measured in terms of how much the split would reduce label impurity. 
An arm $(f,t)$ can be ``pulled'' by computing the reduction of label impurity induced by a new data point sampled from the dataset.
This allows us to compute an estimate $\hat \mu_{ft}$ and a corresponding confidence interval, which can be used in a batched variant of the Upper Confidence Bound (UCB) and successive elimination algorithms~\cite{lai1985asymptotically,zhang2019adaptive} to identify the best arm $(f,t)$. Crucially, \algname uses the adaptive sampling tools of multi-armed bandits to avoid computing the split qualities over the entire dataset.


We demonstrate the benefits of \algname on a variety of datasets, for both classification and regression tasks.
In some settings, \algname algorithm leads to 100x faster training (a 99\% reduction in training time), without any decrease in test accuracy, over an exact implementation of RF that searches for the optimal $(f,t)$ pair via brute-force computation. 
Additionally, we demonstrate similar speedups when using \algname across a variety of forest-based variants, such as Extremely Random Forests and Random Patches. 


\subsection{Related work} \label{sec:related}

Random Forests were originally proposed by Ho~\cite{ho1995random}.
In its original formulation, RF constructs $n_{\text{tree}}$ DTs, where each DT is trained on a bootstrap sample of all $N$ data points and a random subset of the features at each node (a technique known as random subspacing \cite{ho1998random}).
More recently, the need for training RFs on large datasets has prompted the development of several techniques to accelerate training at both the software and hardware levels.


\textbf{Software acceleration of RF:} Most of the software and algorithmic acceleration techniques focus on the training of each individual DT. 
FastForest \cite{yates2021fastforest} accelerates the node-splitting task using three ideas: subsampling a pre-specified number of data points without replacement (subbagging), subsampling a pre-specified number of features dependent on the current number of data points (Dynamic Restricted Subspacing), and dividing values of a given feature into $T$ bins, where $T$ depends on the number of data points at the node (Logarithmic Split-Point Sampling, inspired by the single-tree SPAARC algorithm \cite{yates2018spaarc}).
\algname is inspired by the ideas in FastForest, but does not require the number of data points or features to be prespecified and, instead, determines them by adapting to the data distribution.


Other recent work has also used adaptivity to identify the best split.
For example, Very Fast Decision Trees (VFDTs) \cite{domingos2000mining} and Extremely Fast Decision Trees (EFDTs) \cite{manapragada2018extremely} are incremental decision tree learning algorithms in which trees can be updated in streaming settings.
Acceleration of the node-splitting task is achieved by adaptively selecting a subset of data points sufficient to distinguish the best and second best splits.
These approaches are similar to ours, but the sampled data points are used to evaluate all possible splits. \algnamenospace, in contrast, adaptively discards unpromising splits early.
The F-forest~\cite{fujiwara2019fast} algorithm also applies adaptivity and uses an upper bound on the impurity reduction of each split in order to discard candidate splits.
This is similar in spirit to the goal of \algnamenospace, but is based on a deterministic, conservative upper bound on the impurity reduction (as opposed to \algnamenospace's statistical estimate) and incurs computation linear in the node size for each split, even when considering a fixed number of possible split thresholds per feature.

Other variations of RF have been proposed to improve training time.
Random Patches \cite{louppe2012ensembles} builds trees based on a subset of data points and features that is fixed for each entire forest. 
ExtraTrees (ETs) \cite{geurts2006extremely} draw a random subset of $K$ features at each node and, for each one, chooses a number $R$ of random splits. 
It then selects the split that yields the largest impurity reduction from among these $KR$ candidate splits.

Other recent work attempts to accelerate RF training by identifying the optimal number of decision trees needed in the forest~\cite{oshiro2012many}, a form of hyperparameter tuning. The \algname subroutine can also be incorporated into these methods.

\textbf{Hardware acceleration of RF: }
The training of RFs can also be significantly accelerated through the use of specialized hardware.
For instance, the implementation of RF available in Weka~\cite{eibe2016weka} allows trees to be trained on different cores and reduces forest training time.
A GPU-based parallel implementation of RF has also been proposed in~\cite{grahn2011cudarf}.
These solutions require specialized hardware (e.g., GPU-based PC video cards) and are inappropriate for everyday users locally executing data-mining tasks on standard PCs or smartphones. 
As such, it is still desirable to develop techniques to improve prediction performance and processing speed at an algorithmic, platform-independent level.

\section{Preliminaries \label{sec:prelims}}

We now formally describe the RF algorithm and other tree-based models, all of which rely on a node-splitting subroutine.
We consider $N$ data points $\{(\mathbf{x}_i, y_i)\}_{i=1}^N$ where each $\mathbf{x}_i$ is an $M$-dimensional feature vector and $y_i$ is its target. Following standard literature, we consider flexible feature types such as numerical or categorical. We consider both categorical targets for classification and numerical targets for regression. With a slight abuse of notation, we use $\mathcal{X}$ to mean either the set of indices $\{i\}$ or the values $\{(\mathbf{x}_i, y_i)\}$, with the meaning clear from context.

An RF contains $n_{\text{tree}}$ decision trees, each trained on a set of $N$ bootstrapped data points (sampled with replacement) and a random subset of features at each node. 
The whole RF, an ensemble model, outputs the trees' majority vote in classification and the trees' average prediction in regression \cite{ho1995random,breiman2001random}.
We focus on the top-down approach of constructing DTs by choosing the feature-threshold pair at each step that best splits the set of targets at a given node \cite{rokach2005top}. 

We now discuss the method by which the best split is chosen in each node. Consider a node with $n$ data points $\mathcal{X}$ and $m$ features $\mathcal{F}$. 
Note that $n$ and $m$ at the given node may or may not be the same as $N$ and $M$ of the entire dataset (for example, RF considers a random subset of $m = \sqrt{M}$ features at each node \cite{bernard2009influence}).
Let $\mathcal{X}_{\text{L}, ft}$ and $\mathcal{X}_{\text{R}, ft}$ be the left and right child subsets of $\mathcal{X}$ when $\mathcal{X}$ is split according to feature $f$ at threshold $t$. The approach finds the split that best reduces the label impurity; i.e., finds
\begin{align} \label{eqn:split}
    f^*, t^* = \argmin_{f \in \mathcal{F}, t \in \mathcal{T}_f} \frac{\vert \mathcal{X}_{\text{L}, ft} \vert}{n} I (\mathcal{X}_{\text{L}, ft} ) + \frac{\vert \mathcal{X}_{\text{R}, ft} \vert}{n} I ( \mathcal{X}_{\text{R}, ft} ) - I(\mathcal{X}),
\end{align}
where $I(\mathcal{S})$ measures the impurity of targets $\{y_i\}_{i\in \mathcal{S}}$ and $\mathcal{T}_f$ is the set of allowed thresholds for feature $f$. Common impurity measures include Gini impurity or entropy for classification, and mean-squared-error (MSE) for regression \cite{breiman2017classification}:
\begin{align} \label{eqn:metric}
    \text{Gini}: 1 - \sum_{k=1}^K p_k^2,~~~~ 
    \text{Entropy}: - \sum_{k=1}^K p_k \log_2p_k,~~~~\text{MSE}: \frac{1}{n}\sum_{i \in \mathcal{X}} (y_i - \bar{y})^2,
\end{align}
where $K$ is the total number of classes and $p_k = \frac{1}{n}\sum_{i \in \mathcal{X}} \mathbb{I}(y_i=k)$ is the proportion of class $k$ in $\mathcal{X}$ in classification, and $\bar{y} = \frac{1}{n}\sum_{i \in \mathcal{X}} y_i$ is the average target value in regression. We note that our proposed algorithm, \algnamenospace, does not assume any particular structure of $I(\cdot)$. 

While the conventional RF considers all $(n-1)$ possible splits among the $n$ generally different values in the dataset for a given feature $f$, in this work we focus on the histogram-based variant that chooses the threshold from a set of predefined values $\mathcal{T}_f$, e.g., $\vert \mathcal{T}_f \vert$ equally-spaced histogram bin edges; this variant is substantially more efficient, offers comparable accuracy, and has been used in most state-of-the-art implementations such as XGBoost and LightGBM \cite{ranka1998clouds,chen2016xgboost,ke2017lightgbm,yates2018spaarc}.
A na\"{i}ve algorithm finds the best feature-threshold pairs in Equationh \eqref{eqn:split} by evaluating the label impurity reduction for each feature-threshold over all $n$ data points, which incurs computation linear in $n$.

\section{\algnamenospace}
\label{sec:algo}

We now discuss \algname and how it can reduce the complexity of the node-splitting problem to logarithmic in $n$.
With the same notation as in Section \ref{sec:prelims}, we note that Equation \eqref{eqn:split} is equivalent to 
\begin{align} \label{eqn:split_mab}
    f^*, t^* = \argmin_{f \in \mathcal{F}, t \in \mathcal{T}_f} \frac{\vert \mathcal{X}_{\text{L}, ft} \vert}{n} I (\mathcal{X}_{\text{L}, ft} ) + \frac{\vert \mathcal{X}_{\text{R}, ft} \vert}{n} I ( \mathcal{X}_{\text{R}, ft} )
\end{align}
and so we focus on solving Equation \eqref{eqn:split_mab}.
Let $\mu_{ft} = \frac{\vert \mathcal{X}_{\text{L}, ft} \vert}{n} I (\mathcal{X}_{\text{L}, ft} ) + \frac{\vert \mathcal{X}_{\text{R}, ft} \vert}{n} I ( \mathcal{X}_{\text{R}, ft} )$ be the optimization objective for feature-threshold pair $(f,t)$.
Omitting the dependence on $K$, computing $\mu_{ft}$ exactly is at least $O(n)$.
\algnamenospace, however, \emph{estimates} $\mu_{ft}$ with less computation by drawing $n' < n$ independent samples with replacement from $\mathcal{X}$.
As shown in Subsec. \ref{subsec:CI}, it is possible to construct a point estimate $\hat{\mu}_{ft}(n')$ and a $(1-\delta)$ confidence interval (CI) $C_{ft}(n', \delta)$ for the parameter $\mu_{ft}$, where $n'$ and $\delta$ determine estimation accuracy. 
The width of these CIs generally scales with $\sqrt{\tfrac{\log 1/\delta}{n'}}$.
To estimate the solution to Problem \eqref{eqn:split_mab} with high confidence, we can then choose to sample different amounts of data points for different features so as to estimate their impurity reductions to varying degrees of accuracy. 
Intuitively, promising features-threshold splits with high impurity reductions (lower values of $\mu_{ft}$) should be estimated with high accuracy with many data points, while less promising ones with low impurity reductions (higher values of $\mu_{ft}$) can be discarded early.

The exact adaptive estimation procedure, \algnamenospace, is described in Algorithm \ref{alg:bandit_based_search}. 
It can be viewed as a batched version of the conventional UCB algorithm \cite{lai1985asymptotically,zhang2019adaptive} combined with successive elimination \cite{successiveelimination}, is straightforward to implement, and has been used in other applications \cite{tiwari2020banditpam, tiwari2022fastermips,tiwari2022banditcaching, baharav2022approximate, baharav2019ultra}.
Algorithm \ref{alg:bandit_based_search} uses the set $\mathcal{S}_{\text{solution}}$ to track all potential solutions to Problem \eqref{eqn:split_mab}; $\mathcal{S}_{\text{solution}}$ is initialized as the set of all feature-threshold pairs $\{(f,t)\}$ and Algorithm \ref{alg:bandit_based_search} maintains the mean objective estimate $\hat{\mu}_{ft}$ and $(1-\delta)$ CI $C_{ft}$ for each potential solution $(f,t) \in \mathcal{S}_{\text{solution}}$. 

In each iteration, a new batch of data points $\mathcal{X}_{\text{batch}}$ is used to evaluate the split quality for all potential feature-threshold splits in $\mathcal{S}_{\text{solution}}$, which allows the estimate of each $\hat{\mu}_{ft}$ to be made more accurate. 
Based on the current estimate, if a candidate's lower confidence bound $\hat{\mu}_{ft} - C_{ft}$ is greater than the upper confidence bound of the most promising candidate $\min_{f,t}(\hat{\mu}_{ft} + C_{ft})$, we remove it from  $\mathcal{S}_{\text{solution}}$. This process continues until there is only one candidate in $\mathcal{S}_{\text{solution}}$ or until we have sampled more than $n$ data points. In the latter case, we know that the difference between the remaining candidates in $\mathcal{S}_{\text{solution}}$ is so subtle that an exact computation is warranted.
\algname then compute those candidates' objectives exactly and returns the best candidate in the set. 

\subsection{Point estimates and confidence intervals for impurity metrics \label{subsec:CI}}


We now discuss \algnamenospace's construction of point estimates and confidence intervals of $\mu_{ft}$ based on a set of $n'$ points, $\{(\mathbf{X}_i, Y_i)\}_{i=1}^{n'}$, sampled independently and with replacement from $\mathcal{X}$.
We consider two widely used impurity metrics in classification, Gini impurity and entropy, although mean estimates and confidence intervals for other settings and metrics can be derived similarly (more details are provided in Appendix \ref{appendix:delta}).  

Let $p_{\text{L}, k} \coloneqq \frac{1}{n} \sum_{i=1}^{n} \mathbb{I}(x_{if}<t, y_i=k)$ and $p_{\text{R}, k} \coloneqq \frac{1}{n} \sum_{i=1}^{n} \mathbb{I}(x_{if}\geq t, y_i=k)$ denote the proportion of the full $n$ data points in class $k$ and each of the two subsets created by the split $(f,t)$ (we call these subsets ``left'' and ``right'', respectively). 
Note that
\begin{align}
    &\sum_{k} p_{\text{L}, k} = \frac{|\mathcal{X}_{\text{L}, ft}|}{n} \quad \text{ and } \quad \sum_{k} p_{\text{R}, k} = \frac{|\mathcal{X}_{\text{R}, ft}|}{n}. \label{eq:leftright}
\end{align}
Furthermore, let $\hat{p}_{\text{L}, k} \coloneqq \frac{1}{n'} \sum_{i=1}^{n'} \mathbb{I}(X_{if}<t, Y_i=k)$ and $\hat{p}_{\text{R}, k} \coloneqq \frac{1}{n'} \sum_{i=1}^{n'} \mathbb{I}(X_{if}\geq t, Y_i=k)$ denote the empirical estimates of $p_{\text{L}, k}$ and $p_{\text{R}, k}$  based on the $n'$ subsamples drawn thus far. Then $\{\hat{p}_{\text{L}, k}, \hat{p}_{\text{R}, k}\}_{k=1}^K$ jointly follow a multinomial distribution
satisfying
\begin{align*}
    \mathbb{E}[\hat{p}_{\text{L}, k}] = p_{\text{L}, k},~~~~\text{Var}[\hat{p}_{\text{L}, k}] = \frac{1}{n'} p_{\text{L}, k} (1 - p_{\text{L}, k}), \\
    \mathbb{E}[\hat{p}_{\text{R}, k}] = p_{\text{R}, k},~~~~\text{Var}[\hat{p}_{\text{R}, k}] = \frac{1}{n'} p_{\text{R}, k} (1 - p_{\text{R}, k}).
\end{align*}
since for each random data point $(\mathbf{X}_i, Y_i)$, exactly one element of the set $\{\hat{p}_{\text{L}, k}, \hat{p}_{\text{R}, k}\}_{k=1}^K$ is incremented.
Using Equations~\eqref{eqn:metric} and \eqref{eq:leftright}, and the definition of $\mu_{ft}$ after Equation~\eqref{eqn:split_mab}, we write
\begin{align}
    & \text{Gini impurity}:~\mu_{ft} =  1 - \frac{\sum_{k} p_{\text{L}, k}^2}{\sum_{k} p_{\text{L}, k}} - \frac{\sum_{k} p_{\text{R}, k}^2}{\sum_{k} p_{\text{R}, k}}, \\
    & \text{Entropy}:~\mu_{ft} = - \sum_k p_{\text{L}, k} \log_2 \frac{p_{\text{L}, k}}{\sum_{k'} p_{\text{L}, k'}} - \sum_k p_{\text{R}, k} \log_2 \frac{p_{\text{R}, k}}{\sum_{k'} p_{\text{R}, k'}},
\end{align}
where we can use the empirical parameters $\{\hat{p}_{\text{L}, k}, \hat{p}_{\text{R}, k}\}_{k=1}^K$ as plug-in estimators for the true parameters $\{p_{\text{L}, k}, p_{\text{R}, k}\}_{k=1}^K$ to produce the point estimate $\hat{\mu}_{ft}$.

In \algname (Algorithm~\ref{alg:bandit_based_search}), each batch of $B$ data points is used to update each $\hat{p}_{\text{L}, k}$ and $\hat{p}_{\text{R}, k}$, which are used in turn to update our point estimates $\hat{\mu}_{ft}$ and the corresponding CIs. 
The CIs of $\hat{\mu}_{ft}$ are based on standard error derived using the delta method \cite{van2000asymptotic}.
As in standard applications of the delta method, the estimates $\hat{\mu}_{ft}$ are asymptotically unbiased and their corresponding CIs are asymptotically valid.
Appendix \ref{appendix:delta} provides further details, including a derivation of the CIs and discussion of convergence properties.




\begin{algorithm}[t]

\caption{
\algname (
$\mathcal{X}, 
\mathcal{F}, 
\mathcal{T}_f,
I(\cdot)$,
$B$,
$\delta$
) \label{alg:bandit_based_search}}
\begin{algorithmic}[1]
\State $\mathcal{S}_{\text{solution}} \leftarrow \{(f,t),~ \forall~f \in \mathcal{F},~\forall t \in \mathcal{T}_f\}$ \Comment{Set of potential solutions to Problem \eqref{eqn:split_mab}}
\State $n_{\text{used}} \gets 0$  \Comment{Number of data points sampled}
\State For all $(f,t) \in  \mathcal{S}_{\text{solution}}$, set $\hat{\mu}_{ft} \leftarrow \infty$, $C_{ft} \leftarrow \infty$  \Comment{Initialize mean and CI for each arm}
\ForAll{$f \in \mathcal{F}$}
    \State Create empty histogram, $h_f$, with $|\mathcal{T}_f| = T$ equally spaced bins
\EndFor
\While{$n_{\text{used}} < n $ and $|\mathcal{S}_{\text{solution}}| > 1$} 
        \State Draw a batch sample $\mathcal{X}_{\text{batch}}$ of size $B$ with replacement from $\mathcal{X}$
        \ForAll{unique $f$ in $\mathcal{S}_{\text{solution}}$}
            \ForAll{$x$ in $\mathcal{X}_{\text{batch}}$}
            \State Insert $x_f$ into histogram $h_f$ \Comment{Each insertion is $O(1)$}
            \EndFor
        \EndFor
        \ForAll{$(f,t) \in \mathcal{S}_{\text{solution}} $}
            \State Update $\hat{\mu}_{ft}$ and $C_{ft}$ based on histogram $h_f$ \Comment{For fixed $f$, this is $O(T)$}
        \EndFor
    \State $\mathcal{S}_{\text{solution}} \leftarrow \{(f,t) : \hat{\mu}_{ft} - C_{ft} \leq \min_{f,t}(\hat{\mu}_{ft} + C_{ft})\}$ \Comment{Retain only promising splits}
    \State $n_{\text{used}} \leftarrow n_{\text{used}} + B$
\EndWhile
\If{$\vert \mathcal{S}_{\text{solution}} \vert$ = 1}
    \State \textbf{return} $(f^*,t^*) \in \mathcal{S}_{\text{solution}}$
\Else
    \State Compute $\mu_{ft}$ exactly for all $(f,t) \in \mathcal{S}_{\text{solution}}$
    \State \textbf{return} $(f^*,t^*) = \argmin_{(f,t) \in \mathcal{S}_{\text{solution}}} \mu_{ft}$
\EndIf
\end{algorithmic}
\end{algorithm}

\subsection{Algorithmic and implementation details \label{subsec:algdetails}}
\label{A0}

We considered sampling with replacement in \algname (Algorithm \ref{alg:bandit_based_search}) primarily for the ease of theoretical analysis. In practice, we found sampling without replacement was more computationally efficient and did not significantly change the results, and was used in the actual implementation.

For DTs, in classification tasks, we allow individual DTs to provide soft votes and average their predicted class probabilities to determine the forest's predicted class label, following existing approaches \cite{scikit-learn}. 
This is in contrast with hard votes, in which each DT is only permitted to produce its best label and the forest's prediction is determined by majority voting.
For the fixed-budget experiments in Section \ref{sec:experiments}, we terminate tree construction and do not split nodes further if doing so would violate our budget constraints.

\section{Analysis of the Algorithm \label{sec:theory}}




In this section, we prove that \algname returns the optimal feature-threshold pair for a node split with high probability. 
Furthermore, we provide bounds on computational complexity of \algnamenospace, which can lead to a logarithmic dependence on the number of data points $n$ under weak assumptions.

As above, consider a node with $n$ data points $\mathcal{X}$, $m$ features $\mathcal{F}$, and $T$ possible thresholds for each feature ($\vert \mathcal{T}_f \vert = T$ for all $f$).
Suppose $(f^*, t^*) = \argmin_{f \in \mathcal{F}, t \in \mathcal{T}_f} \mu_{ft}$ is the optimal feature-threshold pair at which to split the node.
For any other feature-threshold pair $(f, t)$,
define $\Delta_{ft} \coloneqq \mu_{ft} - \mu_{f^*t^*}$.
To state the following results, we will assume that, for a fixed feature-threshold pair $(f,t)$ and $n'$ randomly sampled datapoints, the $(1-\delta)$ confidence interval scales as $C_{ft}(n', \delta) = O(\sqrt{\frac{\log 1/\delta}{n'}})$.
(This assumption is justified for the confidence intervals of Gini impurity and entropy under weak assumptions on the $\mu_{f,t}$'s \cite{van2000asymptotic}).
With this assumption, we state the following theorem:

\begin{theorem} \label{thm:specific}
Assume $\exists c_0 > 0$ s.t. $\forall \delta>0, n'>0,$ we have $C_{ft}(n', \delta) < c_0\sqrt{\frac{\log 1/\delta}{n'}}$. 
For $\delta = \frac{1}{n^2 m T}$, with probability at least $1-\tfrac{1}{n}$, Algorithm \ref{alg:bandit_based_search}
returns the correct solution to Equation \eqref{eqn:split_mab}. Furthermore, Algorithm \ref{alg:bandit_based_search}
uses a total of $M$ computations, where
\begin{align} \label{eqn:instance_bd}
\mathbb{E}[M] \leq \sum_{f\in\mathcal{F},t\in \mathcal{T}_f}  \min \left[ \frac{4c_0^2}{\Delta_{ft}^2} \log(n^2 m T) + B, 2n \right] + 2mT.
\end{align}
\end{theorem}

Theorem \ref{thm:specific} is proven in Appendix \ref{appendix:proofs}.
Intuitively, Theorem \ref{thm:specific} states that with high probability, \algname returns the optimal feature-threshold pair at which to split the node.
The bound Equation \eqref{eqn:instance_bd} suggests the computational cost of a feature-threshold pair $(f,t)$, i.e., $\min \left[ \frac{4c_0^2}{\Delta_{ft}^2} \log(n^2 m T) + B, 2n \right]$, depends on $\Delta_{ft}$, which measures how close its optimization parameter $\mu_{ft}$ is to $\mu_{f^*t^*}$. Most reasonably different features $f\neq f^*$ will have a large $\Delta_{ft}$ and incur computational cost $O(\log(n^2 m T))$ that is sublinear in $n$.

In turn, this implies \algname takes only $O(mT\log(n^2mT))$ computations per feature-threshold pair if there is reasonable heterogeneity among them. As proven in Appendix 2 of \cite{bagaria2018medoids}, this is the case under a wide range of distributional assumptions on the $\mu_{ft}$'s, e.g., when the $\mu_{ft}$'s follow a sub-Gaussian distribution across the pairs $(f, t)$.
Such assumptions ensure that \algname has an overall complexity of $O(m T \log (n^2 m T))$, which is sublinear in the number of data points $n$.
We note that in the worst case, however, \algname may take $O(n(m+T))$ computations per feature-threshold pair when most splits are equally good, in which case \algname reduces to a batched version of the na\"ive algorithm.
This may happen, for example, in highly symmetric datasets where all splits reduce the impurity equally.
Other recent work provides further discussion on the conversion between a bound like Equation \eqref{eqn:instance_bd}, which depends on the $\Delta_i$'s, and a bound in terms of other problem parameters such as $O(mT\log(n^2mT))$ under various assumptions on the $\mu_{ft}$'s \cite{bagaria2018medoids,zhang2019adaptive,baharav2019ultra,tiwari2020banditpam,bagaria2021bandit,baharav2022approximate}.

Finally, we note that $\delta$ is a hyperparameter governing the error rate. It is possible to prove results analogous to Theorem \ref{thm:specific} for arbitrary $\delta$.


\section{Experimental Results \label{sec:experiments}}

We demonstrate the advantages of \algname in two settings. In the first setting, the baseline models with and without \algname are trained to completion and we report wall-clock training time and generalization performance. In the second setting, we consider training each forest with a fixed computational budget and study effect of \algname on generalization performance. We provide a description of each dataset in Appendix \ref{appendix:experimentdetails}.



We note that head-to-head wall-clock time comparisons with common implementations of forest-based algorithms, such as Weka's \cite{eibe2016weka} or \texttt{scikit-learn}'s \cite{pedregosa2011scikit}, would be unfair due to their extensive hardware- and language-level optimizations.
As such, we reimplement these baselines in Python and focus on algorithmic improvements.
The only difference between our model and the baselines is the call to the node-splitting subroutine (\algname in our model, and the exact, brute-force solver for the baseline models); thus, any improvements in runtime are due to improvements in the node-splitting algorithm.
This is verified by profiling the implementations and measuring the relative time spent in the node-splitting subroutine versus total runtime (see Appendix \ref{appendix:profiles}).
Our approach allows us to focus on algorithmic improvements as opposed implementation-specific optimizations. Our implementation of these baselines may also be of independent interest and we verify the quality of our implementations, via agreement with \texttt{scikit-learn}, in Appendix \ref{appendix:scikit-learncomparison}. 
We also provide a brief discussion of how an optimized version of our reimplementations may outperform \texttt{scikit-learn} in Appendix \ref{appendix:limitations}. On the MNIST dataset, our optimized implementation trains approximately 4x faster than \texttt{scikit-learn}'s \texttt{DecisionTreeClassifier}.

\paragraph{Baseline Models:} We compare the histogrammed versions of three baselines with and without \algnamenospace: Random Forest (RF), ExtraTrees \cite{geurts2006extremely}, and Random Patches (RP) \cite{louppe2012ensembles}. 
In Random Forests, each tree fits a bootstrap sample of $N$ datapoints and considers random subset of $\sqrt{M}$ features at every node split. 
Extra Trees (also known as Extremely Randomized Forests) are identical to Random Forests except for two differences. 
First, in regression problems, all features are considered at every node split (in classification problems, we still use only $\sqrt{M}$ features). 
Second, the histogram edges of a feature are randomly chosen from a uniform distribution over that feature's minimum and maximum value. 
In classification problems, each histogram has $\sqrt{M}$ bins and in regression problems, each histogram has $M$ bins. 
These conventions follow standard implementations \cite{scikit-learn}. 
Note that in ExtraTrees, the bins in a feature's histogram need not be equally spaced. 
Random Patches is identical to Random Forests but the training dataset is reduced to $\alpha_n$ of its original datapoints and $\alpha_f$ of its original features, where $\alpha_n$ and $\alpha_f$ are prespecified constants, and the subsampled dataset is fixed for the training of the entire forest.
Full settings for all experiments are given in Appendix \ref{appendix:experimentdetails}.

\subsection{Wall-clock time comparisons}

In the first setting, we compare baseline models with and without \algname in terms of wall-clock training time. Tables \ref{table:classificationruntime} and \ref{table:regressionruntime} show that \algname provides similar generalization performance but faster training than the usual na\"ive algorithm for node-splitting for almost all baselines in both classification and regression tasks. Across various tasks, \algname leads to approximately 2x-100x faster training, a reduction of training time of 50-99\%. These benefits are wholly attributable to \algname as the only difference between successive minor rows in each table is the node-splitting subroutine (see Appendix \ref{appendix:profiles} for further discussion).

\begin{table}[]
\resizebox{\textwidth}{!}{
\begin{tabular}{|c|c|c|c|}
\hline
\multicolumn{4}{|c|}{MNIST Dataset ($N = 60,000$)}                                                                                                                                       \\         \specialrule{0.1pt}{0pt}{0pt}

\multicolumn{1}{|c|}{Model}                          & \multicolumn{1}{c|}{Training Time (s)}                  & \multicolumn{1}{c|}{Number of Insertions}           & Test Accuracy                 \\ \hline
\multicolumn{1}{|c|}{RF}                             & \multicolumn{1}{c|}{1542.83 $\pm$ 5.837}         & \multicolumn{1}{c|}{1.44E+08 $\pm$ 4.85E+05}          & 0.777 $\pm$ 0.005          \\
\multicolumn{1}{|c|}{\textbf{RF + MABSplit}}         & \multicolumn{1}{c|}{\textbf{40.359 $\pm$ 0.246}} & \multicolumn{1}{c|}{\textbf{3.37E+06 $\pm$ 1.62E+04}} & \textbf{0.763 $\pm$ 0.008} \\ \hline
\multicolumn{1}{|c|}{ExtraTrees}                     & \multicolumn{1}{c|}{1789.653 $\pm$ 2.396}        & \multicolumn{1}{c|}{1.68E+08 $\pm$ 0.00E+00}          & 0.762 $\pm$ 0.003          \\
\multicolumn{1}{|c|}{\textbf{ExtraTrees + MABSplit}} & \multicolumn{1}{c|}{\textbf{50.217 $\pm$ 0.304}} & \multicolumn{1}{c|}{\textbf{4.32E+06 $\pm$ 7.69E+03}} & \textbf{0.755 $\pm$ 0.002} \\ \hline
\multicolumn{1}{|c|}{RP}                             & \multicolumn{1}{c|}{1421.963 $\pm$ 8.368}        & \multicolumn{1}{c|}{1.32E+08 $\pm$ 6.95E+05}          & 0.771 $\pm$ 0.003          \\
\multicolumn{1}{|c|}{\textbf{RP + MABSplit}}         & \multicolumn{1}{c|}{\textbf{38.415 $\pm$ 0.245}} & \multicolumn{1}{c|}{\textbf{3.17E+06 $\pm$ 1.40E+04}} & \textbf{0.768 $\pm$ 0.003} \\ \hline
\multicolumn{4}{|c|}{APS Failure at Scania Trucks Dataset ($N = 60,000$)}                                                                                                                \\ \specialrule{0.1pt}{0pt}{0pt}
\multicolumn{1}{|c|}{Model}                          & \multicolumn{1}{c|}{Training Time (s)}                  & \multicolumn{1}{c|}{Number of Insertions}           & Test Accuracy                 \\ \hline
\multicolumn{1}{|c|}{RF}                             & \multicolumn{1}{c|}{20.542 $\pm$ 0.048}          & \multicolumn{1}{c|}{3.77E+06 $\pm$ 9.66E+03}          & 0.985 $\pm$ 0.0            \\
\multicolumn{1}{|c|}{\textbf{RF + MABSplit}}         & \multicolumn{1}{c|}{\textbf{0.455 $\pm$ 0.002}}  & \multicolumn{1}{c|}{\textbf{6.94E+04 $\pm$ 2.19E+02}} & \textbf{0.985 $\pm$ 0.0}   \\ \hline
\multicolumn{1}{|c|}{ExtraTrees}                     & \multicolumn{1}{c|}{18.849 $\pm$ 0.027}          & \multicolumn{1}{c|}{3.78E+06 $\pm$ 0.00E+00}          & 0.985 $\pm$ 0.0            \\
\multicolumn{1}{|c|}{\textbf{ExtraTrees + MABSplit}} & \multicolumn{1}{c|}{\textbf{0.406 $\pm$ 0.001}}  & \multicolumn{1}{c|}{\textbf{7.00E+04 $\pm$ 0.00E+00}} & \textbf{0.985 $\pm$ 0.0}   \\ \hline
\multicolumn{1}{|c|}{RP}                             & \multicolumn{1}{c|}{17.63 $\pm$ 0.054}           & \multicolumn{1}{c|}{3.22E+06 $\pm$ 1.18E+04}          & 0.985 $\pm$ 0.0            \\
\multicolumn{1}{|c|}{\textbf{RP + MABSplit}}         & \multicolumn{1}{c|}{\textbf{0.399 $\pm$ 0.003}}  & \multicolumn{1}{c|}{\textbf{5.96E+04 $\pm$ 2.19E+02}} & \textbf{0.985 $\pm$ 0.0}   \\ \hline
\multicolumn{4}{|c|}{Forest Covertype Dataset ($N = 581,012$)}                                                                                                                           \\ \specialrule{0.1pt}{0pt}{0pt}
\multicolumn{1}{|c|}{Model}                          & \multicolumn{1}{c|}{Training Time (s)}                  & \multicolumn{1}{c|}{Number of Insertions}           & Test Accuracy                 \\ \hline
\multicolumn{1}{|c|}{RF}                             & \multicolumn{1}{c|}{117.351 $\pm$ 0.123}         & \multicolumn{1}{c|}{1.86E+07 $\pm$ 0.00E+00}          & 0.559 $\pm$ 0.028          \\
\multicolumn{1}{|c|}{\textbf{RF + MABSplit}}         & \multicolumn{1}{c|}{\textbf{0.88 $\pm$ 0.009}}   & \multicolumn{1}{c|}{\textbf{3.98E+04 $\pm$ 1.79E+02}} & \textbf{0.505 $\pm$ 0.004} \\ \hline
\multicolumn{1}{|c|}{ExtraTrees}                     & \multicolumn{1}{c|}{117.984 $\pm$ 0.119}         & \multicolumn{1}{c|}{1.86E+07 $\pm$ 0.00E+00}          & 0.539 $\pm$ 0.022          \\
\multicolumn{1}{|c|}{\textbf{ExtraTrees + MABSplit}} & \multicolumn{1}{c|}{\textbf{2.942 $\pm$ 0.856}}  & \multicolumn{1}{c|}{\textbf{3.69E+05 $\pm$ 1.22E+05}} & \textbf{0.5 $\pm$ 0.005}   \\ \hline
\multicolumn{1}{|c|}{RP}                             & \multicolumn{1}{c|}{104.456 $\pm$ 0.737}         & \multicolumn{1}{c|}{1.62E+07 $\pm$ 8.31E+04}          & 0.51 $\pm$ 0.008           \\
\multicolumn{1}{|c|}{\textbf{RP + MABSplit}}         & \multicolumn{1}{c|}{\textbf{0.815 $\pm$ 0.004}}  & \multicolumn{1}{c|}{\textbf{3.50E+04 $\pm$ 0.00E+00}} & \textbf{0.507 $\pm$ 0.005} \\ \hline
\end{tabular}
}
\vspace{1pt}
\caption{\label{table:classificationruntime} Wall-clock training time, number of histogram insertions, and test accuracies for various models with and without \algnamenospace. \algname can accelerate these models by over 100x in some cases (an 99\% reduction in training time) while achieving comparable accuracy. The number of histogram insertions correlates strongly with wall-clock training time, which justifies our focus on accelerating the node-splitting algorithm via reductions in sample complexity.}
\end{table}

\begin{table}[]
\centering
\begin{tabular}{|ccc|}
\hline
\multicolumn{3}{|c|}{}                                                                                           \\[-1em]
\multicolumn{3}{|c|}{Beijing Multi-Site Air-Quality Dataset (Regression, $N = 420,768$)}                                                                                               \\ \specialrule{0.1pt}{0pt}{0pt}
\multicolumn{3}{|c|}{}                                                                                           \\[-1em]
\multicolumn{1}{|c|}{Model}                          & \multicolumn{1}{c|}{Training Time (s)}                & Test MSE                     \\ \hline
\multicolumn{1}{|c|}{RF}                             & \multicolumn{1}{c|}{138.782 $\pm$ 1.581}         & 1164.576 $\pm$ 0.761             \\
\multicolumn{1}{|c|}{\textbf{RF + MABSplit}}         & \multicolumn{1}{c|}{\textbf{67.089 $\pm$ 1.682}} & \textbf{1109.542 $\pm$ 23.776}   \\ \hline
\multicolumn{1}{|c|}{ExtraTrees}                     & \multicolumn{1}{c|}{115.592 $\pm$ 3.061}         & 1028.054 $\pm$ 11.355            \\
\multicolumn{1}{|c|}{\textbf{ExtraTrees + MABSplit}} & \multicolumn{1}{c|}{\textbf{53.607 $\pm$ 1.278}} & \textbf{1015.234 $\pm$ 6.535}    \\ \hline
\multicolumn{1}{|c|}{RP}                             & \multicolumn{1}{c|}{108.174 $\pm$ 1.24}          & 1128.299 $\pm$ 25.78             \\
\multicolumn{1}{|c|}{\textbf{RP + MABSplit}}         & \multicolumn{1}{c|}{\textbf{60.639 $\pm$ 3.642}} & \textbf{1125.816 $\pm$ 33.56}    \\ \hline
\multicolumn{3}{|c|}{}                                                                                           \\[-1em]
\multicolumn{3}{|c|}{SGEMM GPU Kernel Performance Dataset (Regression, $N = 241,600$)}                                                                                        \\ \specialrule{0.1pt}{0pt}{0pt} 
\multicolumn{3}{|c|}{}                                                                                           \\[-1em]
\multicolumn{1}{|c|}{Model}                          & \multicolumn{1}{c|}{Training Time (s)}                & Test MSE                     \\ \hline
\multicolumn{1}{|c|}{RF}                             & \multicolumn{1}{c|}{32.606 $\pm$ 0.859}          & 69733.002 $\pm$ 57.401           \\
\multicolumn{1}{|c|}{\textbf{RF + MABSplit}}         & \multicolumn{1}{c|}{\textbf{16.51 $\pm$ 0.224}}  & \textbf{69493.921 $\pm$ 73.133}  \\ \hline
\multicolumn{1}{|c|}{ExtraTrees}                     & \multicolumn{1}{c|}{30.624 $\pm$ 0.686}          & 69734.948 $\pm$ 54.876           \\
\multicolumn{1}{|c|}{\textbf{ExtraTrees + MABSplit}} & \multicolumn{1}{c|}{\textbf{14.086 $\pm$ 0.295}} & \textbf{69585.029 $\pm$ 80.281}  \\ \hline
\multicolumn{1}{|c|}{RP}                             & \multicolumn{1}{c|}{26.091 $\pm$ 0.417}          & 66364.998 $\pm$ 894.568          \\
\multicolumn{1}{|c|}{\textbf{RP + MABSplit}}         & \multicolumn{1}{c|}{\textbf{16.409 $\pm$ 0.952}} & \textbf{66310.138 $\pm$ 896.237} \\ \hline
\end{tabular}
\vspace{1pt}
\caption{\label{table:regressionruntime} Wall-clock training time and test MSEs for various models with and without \algnamenospace. \algname can accelerate these models by up to 2x (an 50\% reduction in training time) while achieving comparable results. We omit the number of histogram insertions in favor of wall-clock time for simplicity; unlike in classification, the different baseline regression models have widely varying histogram bin counts. Since the histogram insertion complexity is different across models, the comparison across models would not be fair.}
\end{table}




\subsection{Fixed budget comparisons}

In the second setting, we consider training models under a fixed computational budget. As before, insertion into a histogram is taken to be an $O(1)$ operation. This is justified if the histogram's thresholds are evenly spaced, wherefore the correct bin in which to insert a value can be indexed into directly.
(If the bins are unevenly spaced, we may perform binary searches to locate the correct bin, which is $O(\log T)$ and does not depend on $n$, or cache the results of these binary searches for an evenly-spaced grid across the range of the given feature's value.)

Intuitively, the \algname algorithm allows for splitting a given node with less data point queries and histogram insertions than the na\"ive solution. As such, when the computational budget is fixed, forests trained with \algname should be able to split more nodes and therefore train more trees than forests trained with the na\"ive solver. Prior work suggests that increasing the number of trees in a forest improves generalization performance by reducing variance at the cost of slightly increased bias \cite{hastie2009statisticallearning}.

Tables \ref{table:classificationbudget} and \ref{table:regressionbudget} demonstrate the generalization performance of different models as the computational budget is held constant for different classification and regression tasks. When using \algnamenospace, the trained forests consist of more trees and demonstrate better generalization performance across all baseline models.

\begin{table}[]
\centering
\begin{tabular}{|ccc|}
\hline
\multicolumn{3}{|c|}{MNIST Dataset ($N = 60,000$)}                                                                             \\ \specialrule{0.1pt}{0pt}{0pt}
\multicolumn{1}{|c|}{Model}                          & \multicolumn{1}{c|}{Number of Trees}         & Test Accuracy                 \\ \hline
\multicolumn{1}{|c|}{RF}                             & \multicolumn{1}{c|}{0.2 $\pm$ 0.179}           & 0.143 $\pm$ 0.026          \\
\multicolumn{1}{|c|}{\textbf{RF + MABSplit}}         & \multicolumn{1}{c|}{\textbf{15.8 $\pm$ 0.179}} & \textbf{0.83 $\pm$ 0.002}  \\ \hline
\multicolumn{1}{|c|}{ExtraTrees}                     & \multicolumn{1}{c|}{0.2 $\pm$ 0.179}           & 0.144 $\pm$ 0.027          \\
\multicolumn{1}{|c|}{\textbf{ExtraTrees + MABSplit}} & \multicolumn{1}{c|}{\textbf{12.0 $\pm$ 0.0}}   & \textbf{0.814 $\pm$ 0.001} \\ \hline
\multicolumn{1}{|c|}{RP}                             & \multicolumn{1}{c|}{1.0 $\pm$ 0.0}             & 0.253 $\pm$ 0.003          \\
\multicolumn{1}{|c|}{\textbf{RP + MABSplit}}         & \multicolumn{1}{c|}{\textbf{16.8 $\pm$ 0.179}} & \textbf{0.832 $\pm$ 0.002} \\ \hline
\multicolumn{3}{|c|}{APS Failure at Scania Trucks Dataset ($N = 60,000$)}                                                                                                      \\ \specialrule{0.1pt}{0pt}{0pt}
\multicolumn{1}{|c|}{Model}                          & \multicolumn{1}{c|}{Number of Trees}         & Test Accuracy                 \\ \hline
\multicolumn{1}{|c|}{RF}                             & \multicolumn{1}{c|}{1.0 $\pm$ 0.0}             & 0.985 $\pm$ 0.0            \\
\multicolumn{1}{|c|}{\textbf{RF + MABSplit}}         & \multicolumn{1}{c|}{\textbf{5.8 $\pm$ 0.179}}  & \textbf{0.989 $\pm$ 0.0}   \\ \hline
\multicolumn{1}{|c|}{ExtraTrees}                     & \multicolumn{1}{c|}{1.0 $\pm$ 0.0}             & 0.985 $\pm$ 0.0            \\
\multicolumn{1}{|c|}{\textbf{ExtraTrees + MABSplit}} & \multicolumn{1}{c|}{\textbf{5.6 $\pm$ 0.219}}  & \textbf{0.989 $\pm$ 0.0}   \\ \hline
\multicolumn{1}{|c|}{RP}                             & \multicolumn{1}{c|}{1.0 $\pm$ 0.0}             & 0.985 $\pm$ 0.0            \\
\multicolumn{1}{|c|}{\textbf{RP + MABSplit}}         & \multicolumn{1}{c|}{\textbf{6.8 $\pm$ 0.179}}  & \textbf{0.989 $\pm$ 0.0}   \\ \hline
\multicolumn{3}{|c|}{Forest Covertype Dataset ($N = 581,012$)}                                                                                         \\ \specialrule{0.1pt}{0pt}{0pt}
\multicolumn{1}{|c|}{Model}                          & \multicolumn{1}{c|}{Number of Trees}         & Test Accuracy                 \\ \hline
\multicolumn{1}{|c|}{RF}                             & \multicolumn{1}{c|}{0.4 $\pm$ 0.219}           & 0.514 $\pm$ 0.019          \\
\multicolumn{1}{|c|}{\textbf{RF + MABSplit}}         & \multicolumn{1}{c|}{\textbf{99.8 $\pm$ 0.179}} & \textbf{0.675 $\pm$ 0.002} \\ \hline
\multicolumn{1}{|c|}{ExtraTrees}                     & \multicolumn{1}{c|}{0.2 $\pm$ 0.179}           & 0.496 $\pm$ 0.006          \\
\multicolumn{1}{|c|}{\textbf{ExtraTrees + MABSplit}} & \multicolumn{1}{c|}{\textbf{23.4 $\pm$ 1.403}} & \textbf{0.677 $\pm$ 0.002} \\ \hline
\multicolumn{1}{|c|}{RP}                             & \multicolumn{1}{c|}{0.6 $\pm$ 0.219}           & 0.534 $\pm$ 0.03           \\
\multicolumn{1}{|c|}{\textbf{RP + MABSplit}}         & \multicolumn{1}{c|}{\textbf{100.0 $\pm$ 0.0}}  & \textbf{0.675 $\pm$ 0.002} \\ \hline
\end{tabular}
\vspace{1pt}
\caption{\label{table:classificationbudget} Classification performance under a fixed computational budget (number of histogram insertions) for various models with and without \algnamenospace. \algname allows for more trees to be trained and leads to better generalization performance.}
\end{table}

%

\begin{table}[]
\centering
\begin{tabular}{|ccc|}
\hline
\multicolumn{3}{|c|}{Beijing Multi-Site Air-Quality Dataset ($N = 420,768$)}                                                                                                             \\ \specialrule{0.1pt}{0pt}{0pt}
\multicolumn{1}{|c|}{Model}                          & \multicolumn{1}{c|}{Number of Trees}        & Test MSE                         \\ \hline
\multicolumn{1}{|c|}{RF}                             & \multicolumn{1}{c|}{0.0 $\pm$ 0.0}            & 3208.93 $\pm$ 0.0                  \\
\multicolumn{1}{|c|}{\textbf{RF + MABSplit}}         & \multicolumn{1}{c|}{\textbf{12.0 $\pm$ 0.0}}  & \textbf{927.013 $\pm$ 2.042}       \\ \hline
\multicolumn{1}{|c|}{RP}                             & \multicolumn{1}{c|}{0.0 $\pm$ 0.0}            & 3208.93 $\pm$ 0.0                  \\
\multicolumn{1}{|c|}{\textbf{RP + MABSplit}}         & \multicolumn{1}{c|}{\textbf{11.0 $\pm$ 0.4}}  & \textbf{875.764 $\pm$ 3.064}       \\ \hline
\multicolumn{1}{|c|}{ExtraTrees}                     & \multicolumn{1}{c|}{0.0 $\pm$ 0.0}            & 3208.93 $\pm$ 0.0                  \\
\multicolumn{1}{|c|}{\textbf{ExtraTrees + MABSplit}} & \multicolumn{1}{c|}{\textbf{9.0 $\pm$ 0.0}}   & \textbf{834.338 $\pm$ 4.377}       \\ \hline
\multicolumn{3}{|c|}{SGEMM GPU Kernel Performance Dataset ($N = 241,600$)}                                                                                                             \\ \specialrule{0.1pt}{0pt}{0pt}
\multicolumn{1}{|c|}{Model}                          & \multicolumn{1}{c|}{Number of Trees}        & Test MSE                         \\ \hline
\multicolumn{1}{|c|}{RF}                             & \multicolumn{1}{c|}{0.0 $\pm$ 0.0}            & 131323.839 $\pm$ 0.0               \\
\multicolumn{1}{|c|}{\textbf{RF + MABSplit}}         & \multicolumn{1}{c|}{\textbf{5.6 $\pm$ 0.219}} & \textbf{28571.393 $\pm$ 357.433}   \\ \hline
\multicolumn{1}{|c|}{RP}                             & \multicolumn{1}{c|}{0.8 $\pm$ 0.179}          & 102616.047 $\pm$ 6647.02           \\
\multicolumn{1}{|c|}{\textbf{RP + MABSplit}}         & \multicolumn{1}{c|}{\textbf{2.8 $\pm$ 0.593}} & \textbf{64876.329 $\pm$ 13350.921} \\ \hline
\multicolumn{1}{|c|}{ExtraTrees}                     & \multicolumn{1}{c|}{0.0 $\pm$ 0.0}            & 131323.839 $\pm$ 0.0               \\
\multicolumn{1}{|c|}{\textbf{ExtraTrees + MABSplit}} & \multicolumn{1}{c|}{\textbf{5.0 $\pm$ 0.0}}   & \textbf{29919.254 $\pm$ 344.409}   \\ \hline
\end{tabular}
\vspace{1pt}
\caption{\label{table:regressionbudget} Regression performance under a fixed computational budget (number of histogram insertions). \algname allows for more trees to be trained and leads to better generalization performance.}
\end{table}



\subsection{Feature stability comparisons}

We also apply \algname to compute feature importances under a fixed budget. We follow the common approach of computing the feature importances of multiple forests and then measuring the stability of feature selection across forests using Permutation Feature Importance and Mean Decrease in Impurity (MDI) \cite{nicodemus2011featurestability, piles2021piggenes} (see Appendix \ref{appendix:experimentdetails} for a further discussion of these metrics). The forests trained with \algname demonstrate better feature stabilities than those trained with the na\"ive algorithm; see Table \ref{table:featureimportance}. Note that the datasets used for these experiments are different from the real-world datasets used in the other experiments and are described in Appendix \ref{appendix:experimentdetails}.

\begin{table}[]
\resizebox{\textwidth}{!}{
\begin{tabular}{|c|c|c|c|}
\hline
Importance Model       & Stability Metric    & Dataset                        & Stability                \\ \hline
RF                     & MDI                  & Random Classification          & 0.536 $\pm$ 0.039          \\
\textbf{RF + MABSplit} & \textbf{MDI}         & \textbf{Random Classification} & \textbf{0.863 $\pm$ 0.016} \\ \hline
RF                     & MDI                  & Random Regression              & 0.134 $\pm$ 0.021          \\
\textbf{RF + MABSplit} & \textbf{MDI}         & \textbf{Random Regression}     & \textbf{0.674 $\pm$ 0.043} \\ \hline
RF                     & Permutation          & Random Classification          & 0.579 $\pm$ 0.023          \\
\textbf{RF + MABSplit} & \textbf{Permutation} & \textbf{Random Classification} & \textbf{0.69 $\pm$ 0.023}  \\ \hline
RF                     & Permutation          & Random Regression              & 0.116 $\pm$ 0.017          \\
\textbf{RF + MABSplit} & \textbf{Permutation} & \textbf{Random Regression}     & \textbf{0.437 $\pm$ 0.044} \\ \hline
\end{tabular}
}
\vspace{1pt}
\caption{\label{table:featureimportance} Stability scores under a fixed computational budget (number of histogram insertions). \algname allows more trees to be trained, which leads to greater feature stabilities across the forests.}
\end{table}




%



\section{Discussions and Conclusions \label{sec:discussion}}

In this work, we presented a novel algorithm, \algnamenospace, for determining the optimal feature and corresponding threshold at which to split a node in tree-based learning models.
Unlike prior models such as Random Patches, in which the subsampling hyperparameters $\alpha_n$ and $\alpha_f$ must be prespecified manually, \algname requires no tuning and queries only as much data as is needed by virtue of its adaptivity to the data distribution.
Indeed, robustness to choice of hyperparameters is one of primary appeals of algorithms like RF \cite{probst2019hyperparameters}.

\algname avoids the expensive $O(n\text{log}n)$ sort used in many existing baselines and the $O(n)$ computational complexity of their corresponding histogrammed versions.
\algname can be used in conjunction with existing software- and hardware-specific optimizations and with other methods such as Logarithmic Split-Point Sampling and boosting \cite{yates2021fastforest}.
In boosting, \algname has the potential advantage of only needing to update data points' targets on-the-fly, as needed by its sampling, as opposed to current approaches that update targets for the entire dataset at each iteration.
Additionally, \algname may permit easier parallelization due to lower memory requirements than existing algorithms, which may enable greater use in edge computing and may be adaptable to streaming settings.

We also note that log$n^2$ term in the complexity of \algname (Equation \ref{eqn:instance_bd}) comes from a union bound over all possible $n^2$ confidence intervals computed for each feature-threshold pair across all stages of the algorithm (see the proof of Theorem \ref{thm:specific} in Appendix \ref{appendix:proofs}). This union bound may be weak. Instead, one can more precisely compute the number of confidence intervals in terms of the arm gaps (the $\Delta_i$'s). In this way, the complexity result (Equation \ref{eqn:instance_bd}) can be phrased in terms of the $\Delta_i$'s instead of $n$. Intuitively, this would lead to a complexity bound in terms of the data-generating distribution that is independent of dataset size. Indeed, this is how complexity results are usually stated for multi-armed bandit algorithms \cite{jamieson2014best}. We leave a more detailed investigation of this topic to future work.

\textbf{Acknowledgements:} M. T. was funded by a J.P. Morgan AI Fellowship, a Stanford Indisciplinary Graduate Fellowship, a Stanford Data Science Scholarship, and an Oak Ridge Institute for Science and Engineering Fellowship. M.J.Z. is supported by NIH grant R01 MH115676. I. S. was supported in part by the National Science Foundation (NSF) under grant CCF-2046991.











\bibliographystyle{plain}
\bibliography{refs}

\section*{Checklist}

\begin{enumerate}

\item For all authors...
\begin{enumerate}
  \item Do the main claims made in the abstract and introduction accurately reflect the paper's contributions and scope?
    \answerYes{We have all reviewed the claims in the abstract and introduction and confirmed that they reflect the paper's contributions and scope.}
  \item Did you describe the limitations of your work?
    \answerYes{We discuss the assumptions necessary for \algname to work in Section \ref{sec:theory} and also discuss failure modes when these assumptions are violated. We also describe further room to improve our work in Section \ref{sec:discussion}}
  \item Did you discuss any potential negative societal impacts of your work?
    \answerYes{We discuss potential negative societal impacts in the Future Work section}
  \item Have you read the ethics review guidelines and ensured that your paper conforms to them?
    \answerYes{We have all reviewed the ethics review guidelines and ensured that our paper conforms to them.}
\end{enumerate}

\item If you are including theoretical results...
\begin{enumerate}
  \item Did you state the full set of assumptions of all theoretical results?
    \answerYes{The statement of Theorem \ref{thm:specific} contains a full description of assumptions, which is further expounded upon before and after the theorem statement.}
        \item Did you include complete proofs of all theoretical results?
    \answerYes{The proof of Theorem \ref{thm:specific} is provided in Appendix \ref{appendix:proofs}}
\end{enumerate}

\item If you ran experiments...
\begin{enumerate}
  \item Did you include the code, data, and instructions needed to reproduce the main experimental results (either in the supplemental material or as a URL)?
    \answerYes{Yes, full specifications of all experiments are provided in Section \ref{sec:experiments} and Appendix \ref{appendix:experimentdetails}, and full, documented source code is provided to reproduce our results.}
  \item Did you specify all the training details (e.g., data splits, hyperparameters, how they were chosen)?
    \answerYes{Yes, full specifications of all experiments are provided in Section \ref{sec:experiments} and Appendix \ref{appendix:experimentdetails}, and full, documented source code is provided to reproduce our results.}
\item Did you report error bars (e.g., with respect to the random seed after running experiments multiple times)?
    \answerYes{Every numerical measurement in Tables \ref{table:classificationruntime}, \ref{table:regressionruntime}, \ref{table:classificationbudget}, and \ref{table:regressionbudget} contains confidence intervals}
        \item Did you include the total amount of compute and the type of resources used (e.g., type of GPUs, internal cluster, or cloud provider)?
    \answerYes{Full specifications of computing resources are provided in \ref{appendix:experimentdetails}}
\end{enumerate}

\item If you are using existing assets (e.g., code, data, models) or curating/releasing new assets...
\begin{enumerate}
  \item If your work uses existing assets, did you cite the creators?
    \answerYes{Yes, the only existing assets we used were publicly available datasets, which we cited in Section \ref{subsec:data}}
  \item Did you mention the license of the assets?
    \answerYes{We explicitly mention the licenses in Section \ref{sec:experiments}.}
  \item Did you include any new assets either in the supplemental material or as a URL?
    \answerYes{Yes, the only new assets created are in the form of code, which is entirely submitted in the supplemental material}
  \item Did you discuss whether and how consent was obtained from people whose data you're using/curating?
    \answerYes{Yes, both datasets we use are publicly available and we mention the licenses in Section \ref{sec:experiments}.}
  \item Did you discuss whether the data you are using/curating contains personally identifiable information or offensive content?
    \answerNA{The two datasets are the MNIST dataset and a mathematical dataset; neither have any personally identifiable information or offensive content.}
\end{enumerate}

\item If you used crowdsourcing or conducted research with human subjects...
\begin{enumerate}
  \item Did you include the full text of instructions given to participants and screenshots, if applicable?
    \answerNA{We did not use crowdsourcing or conduct research with human subjects.}
  \item Did you describe any potential participant risks, with links to Institutional Review Board (IRB) approvals, if applicable?
    \answerNA{We did not use crowdsourcing or conduct research with human subjects.}
  \item Did you include the estimated hourly wage paid to participants and the total amount spent on participant compensation?
    \answerNA{We did not use crowdsourcing or conduct research with human subjects.}
\end{enumerate}

\end{enumerate}



\clearpage
\section*{Appendix}
\label{appendix}

\renewcommand\thefigure{\arabic{figure}}
\renewcommand{\figurename}{Appendix Figure}
\renewcommand{\tablename}{Appendix Table}

\setcounter{figure}{0}      
\setcounter{table}{0}      
\setcounter{section}{0}

\section{Proofs}
\label{appendix:proofs}

In this section, we present the proof of Theorem \ref{thm:specific}. 

\begin{proof}
Following the multi-armed bandit literature, we refer to each feature-threshold pair $(f, t)$ as an arm and refer to its optimization objective $\mu_{ft}$ as the arm parameter. Pulling an arm corresponds to evaluating the change in impurity induced by one data point at one feature-threshold pair $(f,t)$ (i.e., arm) and incurs an $O(1)$ computation. This allows us to focus on the number of arm pulls, which translates directly to sample complexity.

First, we show that, with probability at least $1-\tfrac{1}{n}$, all confidence intervals computed throughout the algorithm are valid, in that they contain the true parameter $\mu_{ft}$.
For a fixed $(f,t)$ and a given iteration of the algorithm, the $(1-\delta)$ confidence interval satisfies 
\begin{align*}
    \Pr\left( \left| \mu_{ft} - \hat \mu_{ft} \right| > C_{ft} \right) \leq \delta.
\end{align*}
Let $B$ denote the batch size chosen for \algnamenospace. Note that there are at most $\frac{n}{B}$ rounds in the main \texttt{while} loop (Line 6) of Algorithm \ref{alg:bandit_based_search} and hence at most 
$\frac{nmT}{B} \leq nmT$ confidence intervals computed across all arms and all steps of the algorithm.
With $\delta = \frac{1}{n^2 m T}$, we see that $\mu_{ft} \in [\hat \mu_{ft} - C_{ft}, \hat \mu_{ft} + C_{ft}]$ for every arm $(f,t)$ and for every step of the algorithm with probability at least $1-\frac{1}{n}$, by a union bound over at most $n m T$ confidence intervals. 


Next, we prove the correctness of Algorithm \ref{alg:bandit_based_search}.
Let $(f^*, t^*) = \argmin_{f \in \mathcal{F}, t \in \mathcal{T}_f} \mu_{ft}$ be the desired output of the algorithm.
Since the main \texttt{while} loop in the algorithm can only run $\frac{n}{B}$ times, the algorithm must terminate.
Furthermore, if all confidence intervals throughout the algorithm are correct, it is impossible for $(f^*, t^*)$ to be removed from the set of candidate arms. 
Hence, $(f^*, t^*)$ (or some $(f,t)$ with $\mu_{ft} = \mu_{f^*t^*}$) must be returned upon termination with probability at least $1-\frac{1}{n}$. This proves the correctness of Algorithm \ref{alg:bandit_based_search}.

Finally, we consider the complexity of Algorithm \ref{alg:bandit_based_search}. 
Let $n_{\text{used}}$ be the total number of arm pulls computed for each arm remaining in the set of candidate arms at a given point in the algorithm.
Notice that, for any suboptimal arm $(f,t) \ne (f^*,t^*)$ that has not left the set of candidate arms, we must have
$C_{ft} \leq c_0 \sqrt{ \frac{\log 1/\delta}{n_{\text{used}}}}$ by assumption.
With $\delta = \frac{1}{n^2 m T}$ as above and $\Delta_{ft} = \mu_{ft} - \mu_{f^*t^*}$, if $n_{\text{used}} > \frac{4c_0^2}{\Delta_{ft}^2} \log(n^2 m T)$ then
\aln{
2(C_{ft} + C_{f^*t^*}) \leq 2 c_0 \sqrt{  { \log(n^2 m T) } / {n_{\text{used}}  }} < \Delta_{ft} = \mu_{ft} - \mu_{f^*t^*},
}
and
\begin{align*}
    \hat \mu_{ft} - C_{ft} &> \mu_{ft} - 2C_{ft} \\
    &= \mu_{f^*t^*} + \Delta_{ft} - 2C_{ft} \\
    &\geq \mu_{f^*t^*} + 2 C_{f^*t^*} \\
    &> \hat \mu_{f^*t^*} + C_{f^*t^*}
\end{align*}

which means that $(f,t)$ must be removed from the set of candidate arms at the end of that iteration.
Hence, the number of data point computations $M_{ft}$ required for any arm $(f,t) \ne (f^*, t^*)$ is at most
\aln{
M_{ft} \leq \min \left[ \frac{4c_0^2}{\Delta_{ft}^2} \log(n^2 m T) + B, 2n \right].
}
Notice that this holds simultaneously for all arms $(f,t)$ with probability at least $1-\tfrac{1}{n}$.
We conclude that the total number of arm pulls $M$ satisfies
\aln{
\mathbb{E}[M] & \leq \mathbb{E}[M | \text{ all confidence intervals are correct}] + \frac{1}{n} (2nMT) \\
& \leq \sum_{f\in\mathcal{F},t\in \mathcal{T}_f}  \min \left[ \frac{4c_0^2}{\Delta_{ft}^2} \log(n^2 m T) + B, 2n \right] + 2mT,
}
where we used the fact that the maximum number of computations for any arm is $2n$. As argued before, since each arm pull involves an $O(1)$ computation, $M$ also corresponds the total number of computations.
\end{proof}

\clearpage
\section{$O(\log n)$ scaling of \algname}
\label{appendix:scaling}

In Theorem \ref{thm:specific}, we demonstrated that \algname scales logarithmically in dataset size. In this section, we empirically validate this claim.

\begin{center}
\begin{figure}[ht!]
    \begin{subfigure}{.95\textwidth}
        \centering
        \includegraphics[width=\linewidth]{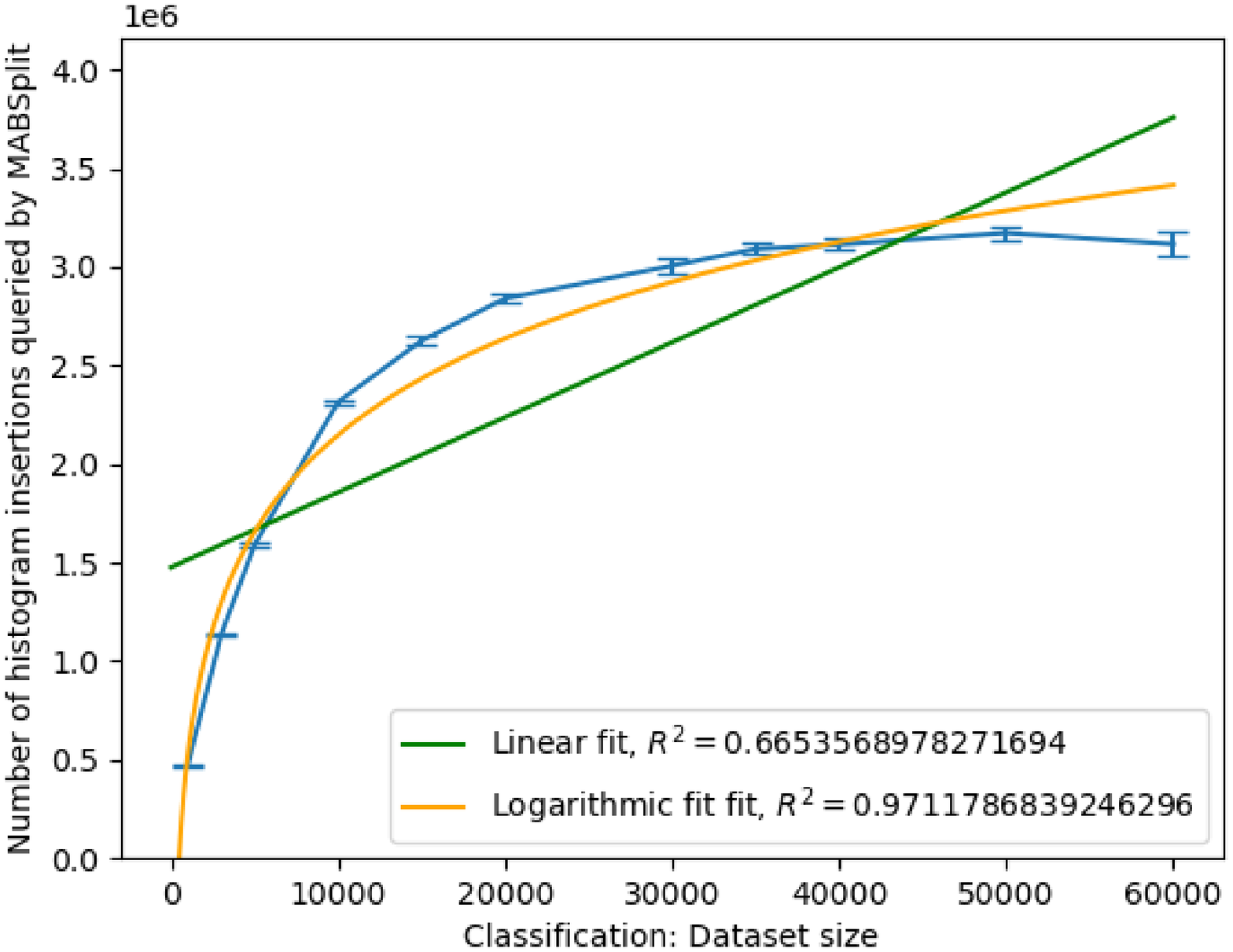} 
        \caption{}
    \end{subfigure}
    \begin{subfigure}{.95\textwidth}
      \centering
      \includegraphics[width=\linewidth]{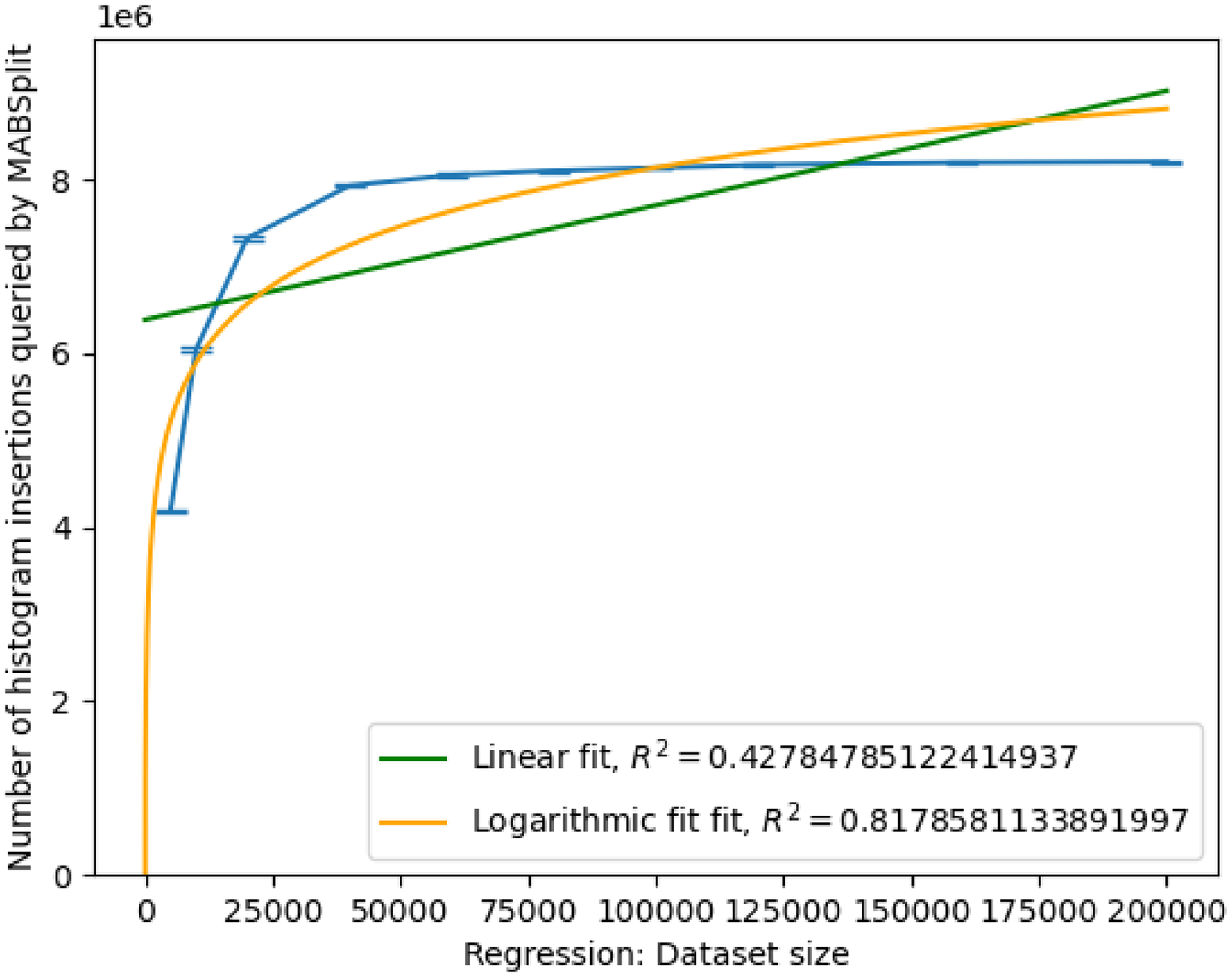}   
      \caption{}
    \end{subfigure}
    \caption{Scaling of MABSplit with different data subset sizes for (a) the MNIST digit classification task and (b) the Random Linear Model regession task. In both tasks, MABSplit appears to scale logarithmically, not linearly, with dataset size.}
    \label{fig:roundcounts}
\end{figure}
\end{center}

Appendix Figure \ref{fig:roundcounts} (a) demonstrates the number of data points queried by \algname for a single node split, i.e., a single call to \algnamenospace, as the dataset size increases, for various subset sizes of MNIST.
For each sample size, a sample is drawn with replacement from the original MNIST dataset. 
The model is trained using Gini impurity in for the usual digitr classification task.

Appendix Figure \ref{fig:roundcounts} (b) demonstrates the same plot for various subset sizes of the Random Linear Model dataset.
The Random Linear Model dataset consists of 200,000 datapoints with 50 features, 6 of which are correlated with the targets and 44 of which are pure noise, using \texttt{scikit-learn}'s \texttt{make\_regression} function \cite{scikit-learn}; 160,000 datapoints are used for training and the remaining 40,000 for test.

Appendix Figure \ref{fig:roundcounts} also shows the best linear and logarithmic fits to each dataset. The relatively high $R^2$ values of the logarithmic fits ($R^2 = 0.97$ and $R^2 = 0.82$) compared to that of the linear fits ($R^2 = 0.66$ and $R^2 = 0.43$) suggests that the scaling of \algname is logarithmic (and therefore sublinear) with dataset size.


\clearpage
\section{Mean Estimation and Confidence Interval Constructions \label{appendix:delta}
}

In this section, we discuss the estimation of the means $\mu_{ft}$ and construction of their confidence intervals via plug-in estimators and the delta method.

Let $p_{\text{L}, k}$, $p_{\text{R}, k}$, $\hat{p}_{\text{L}, k}$, and $\hat{p}_{\text{R}, k}$ be the same as defined in Subsection \ref{subsec:CI}. Furthermore, let $\mathbf{p} = [p_{\text{L}, 1}, \cdots, p_{\text{L}, K}, p_{\text{R}, 1}, \cdots, p_{\text{R}, K}]^T$ and $\hat{\mathbf{p}} = [\hat{p}_{\text{L}, 1}, \cdots, \hat{p}_{\text{L}, K}, \hat{p}_{\text{R}, 1}, \cdots, \hat{p}_{\text{R}, K}]^T$. Then, $n'\hat{\mathbf{p}}$ follows a multinomial distribution with parameters $(n', 2K, \mathbf{p})$. 

Let $\boldsymbol{\theta} = [p_{\text{L}, 1}, \cdots, p_{\text{L}, K}, p_{\text{R}, 1}, \cdots, p_{\text{R}, K-1}]^T$ and $\hat{\boldsymbol{\theta}} = [\hat{p}_{\text{L}, 1}, \cdots, \hat{p}_{\text{L}, K-1}, \hat{p}_{\text{R}, 1}, \cdots, \hat{p}_{\text{R}, K-1}]^T$. Then, by the Central Limit Theorem,
\begin{align}
    \sqrt{n'}(\hat{\boldsymbol{\theta}} - \boldsymbol{\theta}) \overset{D}{\sim} \mathcal{N}(0, \Sigma),
\end{align}
where $\Sigma_{ii} = \theta_i (1-\theta_i)$ and $\Sigma_{ij}=-\theta_i \theta_j$.

Next, we write $\mu_{ft}$ in terms of $\boldsymbol{\theta}$ for the impurity metrics as 
\begin{align}
    \text{Gini impurity}:~ & \mu_{ft}(\boldsymbol{\theta}) =  1 - \frac{\sum_{k=1}^{K} \theta_k^2}{\sum_{k=1}^{K} \theta_k} - \frac{\sum_{k=K+1}^{2K-1} \theta_k^2 + (1 - \sum_{k=1}^{2K-1} \theta_k)^2}{1 - \sum_{k=1}^{K} \theta_k}, \\
    \text{Entropy}:~ & \mu_{ft}(\boldsymbol{\theta}) = - \sum_{k=1}^K \theta_k \log_2 \frac{\theta_k}{\sum_{k'=1}^{K} \theta_k'} - \sum_{k=K+1}^{2K-1} \theta_k \log_2 \frac{\theta_k}{1-\sum_{k'=1}^{K} \theta_k'} - \notag \\
    & (1 - \sum_{k=1}^{2K-1} \theta_k) \log_2 \frac{(1 - \sum_{k=1}^{2K-1} \theta_k)}{1-\sum_{k=1}^{K} \theta_k}.
\end{align}

For a given impurity metric, let $\nabla \mu_{ft}(\boldsymbol{\theta})$ be the derivative of $\mu_{ft}$ with respect to $\boldsymbol{\theta}$. From the delta method, 
\begin{align}
    \sqrt{n'}( \hat{\mu}_{ft}(\boldsymbol{\theta}) - \mu_{ft}(\boldsymbol{\theta})) \overset{D}{\sim} \mathcal{N}(0, \nabla \mu_{ft}(\boldsymbol{\theta})^T \Sigma \nabla \mu_{ft}(\boldsymbol{\theta})),
\end{align}
where the CIs can be constructed accordingly. These CIs are asymptotically valid as $n',n \rightarrow \infty$. For other impurity metrics such as MSE, the CIs can be similarly derived by writing the corresponding $\mu_{ft}$ in terms of $\boldsymbol{\theta}$ and computing $\nabla \mu_{ft}(\boldsymbol{\theta})$.

\clearpage
\section{Comparison of baseline implementations and scikit-learn}
\label{appendix:scikit-learncomparison}

In this section, we compare our re-implementation of common baselines to those in popular packages to verify the accuracy of our re-implementation. Specifically, we compare our implementations of Random Forest Classifiers, Random Forest Regressors, Extremely Random Forest Classifiers, and Extremely Random Forest Regressors to those of $\texttt{scikit-learn}$. We omit comparisons of the Random Patches models because their correctness is implied by that of the Random Forest model, as the Random Patches model consists of applying the Random Forest model to subsampled data and features.

For classification, we compare our implementations on the 20 newsgroups dataset filtered to two newsgroups, $\texttt{alt.atheism}$ and $\texttt{sci.space}$. The dataset is embedded via TF-IDF and projected onto their top 100 principal components, following standard practice \cite{scikit-learn}. The train-test split is the standard one provided by \texttt{scikit-learn}.

For all classification problems, we average the predicted probabilities of each tree in the forest ("soft voting") as opposed to only allowing each tree to vote for a single class ("hard voting"), following the implementation in \texttt{scikit-learn} \cite{scikit-learn}.

For regression, we compare our implementations on the California Housing dataset, subsampled to 1,000 points as performing the regression on the full dataset of approximately 20,000 points is computationally prohibitive. The train-test split is the standard one provided by \texttt{scikit-learn} \cite{scikit-learn}.

Table \ref{table:scikit-learncomparison} presents our results. In all cases, our re-implemented baselines do not present a statistically significant difference in performance from the models present in $\texttt{scikit-learn}$, which suggests that our re-implementations are correct. Performance is measured over 20 random seeds to compute averages and standard deviations.

\begin{table}
\begin{center}
{\begin{tabular}{
|>{\centering\arraybackslash}m{2.0cm}
|>{\centering\arraybackslash}m{2.2cm}
|>{\centering\arraybackslash}m{1.4cm}
|>{\centering\arraybackslash}m{2.5cm}|}
\hline
Model & Task and Dataset & Performance Metric & Test Performance \\
\hline
RF (ours) & Classification: 20 Newsgroups & Accuracy & 74.1 $\pm$ 2.8\% \\
RF ($\texttt{scikit-learn}$) & Classification: 20 Newsgroups & Accuracy & 76.2 $\pm$ 1.7\% \\

\hline
ExtraTrees (ours) & Classification: 20 Newsgroups & Accuracy & 66.5 $\pm$ 5.1\% \\
ExtraTrees ($\texttt{scikit-learn}$) & Classification: 20 Newsgroups & Accuracy & 62.6 $\pm$ 2.8\% \\
\hline
RF (ours) & Regression: California Housing & MSE & 0.679 $\pm$ 0.022 \\
RF ($\texttt{scikit-learn}$) & Regression: California Housing & MSE & 0.672 $\pm$ 0.028 \\
\hline
ExtraTrees (ours) & Regression: California Housing & MSE & 0.696 $\pm$ 0.055 \\
ExtraTrees ($\texttt{scikit-learn}$) & Regression: California Housing & MSE & 0.695 $\pm$ 0.082 \\
\hline
\end{tabular}}
\end{center}
\caption{\label{table:scikit-learncomparison} Comparison of our re-implementation of baselines with the the implementations available in $\texttt{scikit-learn}$. No statistically significant differences are apparent, which suggests that our re-implementations are accurate. }
\end{table}

\clearpage
\section{Profiles}
\label{appendix:profiles}

In this work, we focused on the reducing the runtime at the \textit{algorithmic} level, i.e., reducing the complexity of computing the best feature-threshold split. In this section, we justify this choice by demonstrating that most of the time spent in our re-implementation of the baseline algorithms is spent in computing the best feature-threshold split.

\begin{figure}[h]
    \centering
    \begin{subfigure}{.9\textwidth}
      \centering
      \includegraphics[width=\linewidth]{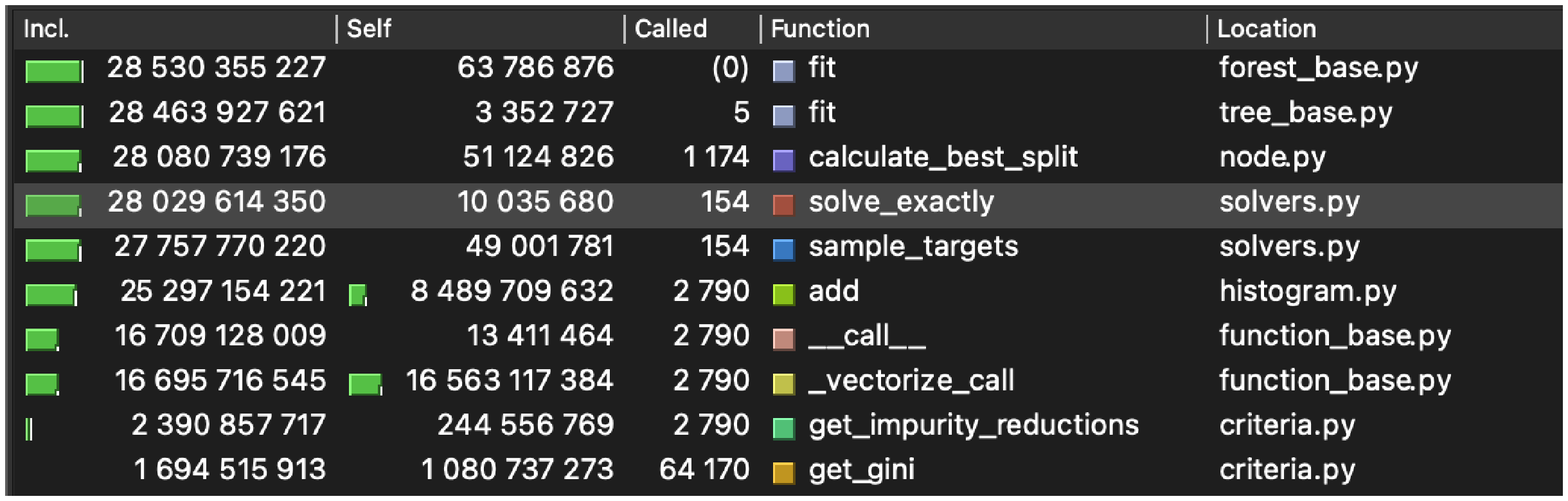}  
      \caption{}
    \end{subfigure}
    \begin{subfigure}{.9\textwidth}
      \centering
      \includegraphics[width=\linewidth]{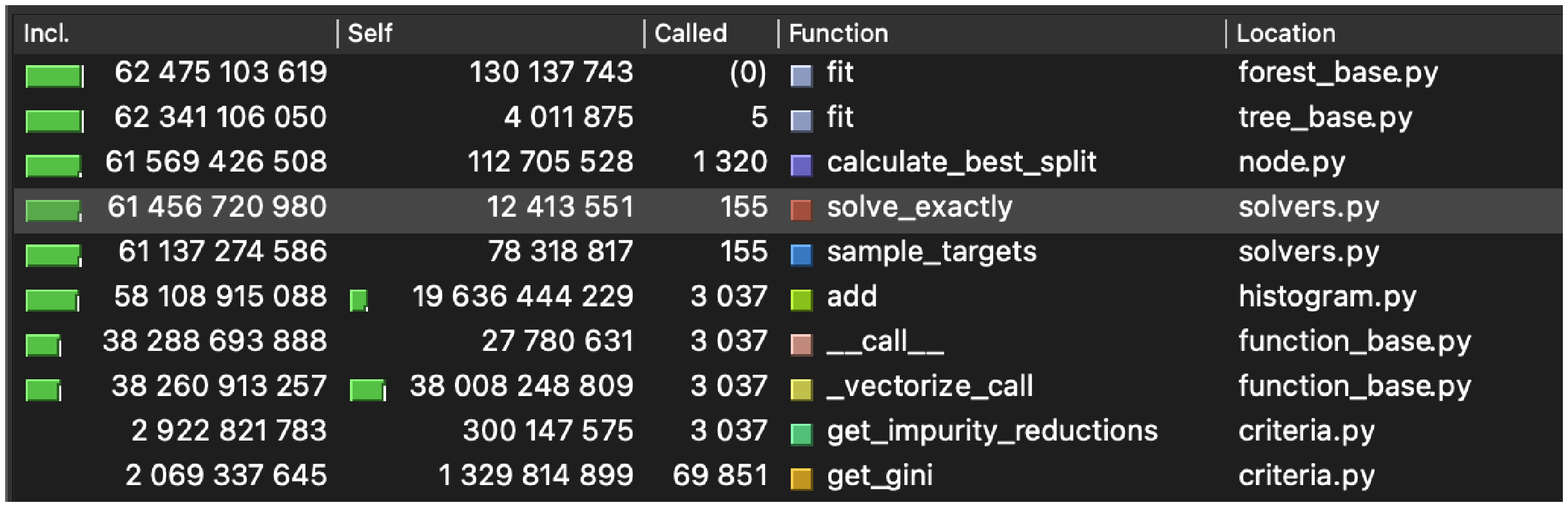}   
      \caption{}
    \end{subfigure}
    \caption{Profiles for the node-splitting algorithm using the exact solver/na\"ive computation, the canonical algorithm for computing the best feature-threshold split, for 5,000 (top) and 10,000 (bottom) data point subsets of MNIST. The "Function" column is the name of the called function, the "Incl." column is the time spent in the function and any called subroutines, and the "Self" column is the time (in nanoseconds) spent in only the function and \textit{not} in any callees. All times are in nanoseconds. When increasing the dataset size, the overhead spent outside of the $\texttt{solve\_exactly}$ function grows negligibly from about 0.5 seconds to about 1 second. However, the time spent in the $\texttt{solve\_exactly}$ function and any called subroutines grows from about 28 seconds to about 61 seconds and constitutes approximately 98\% of the increase in wall-clock time. This observation motivates our focus on improving the subroutine used to identify the best feature-threshold split. This profile was generated with $\texttt{cProfile}$ and visualized with $\texttt{pyprof2calltree}$ \cite{lanaro2019python}.}
    \label{fig:theirprofiles}
\end{figure}

\begin{figure}[h]
    \centering
    \begin{subfigure}{.9\textwidth}
      \centering
      \includegraphics[width=\linewidth]{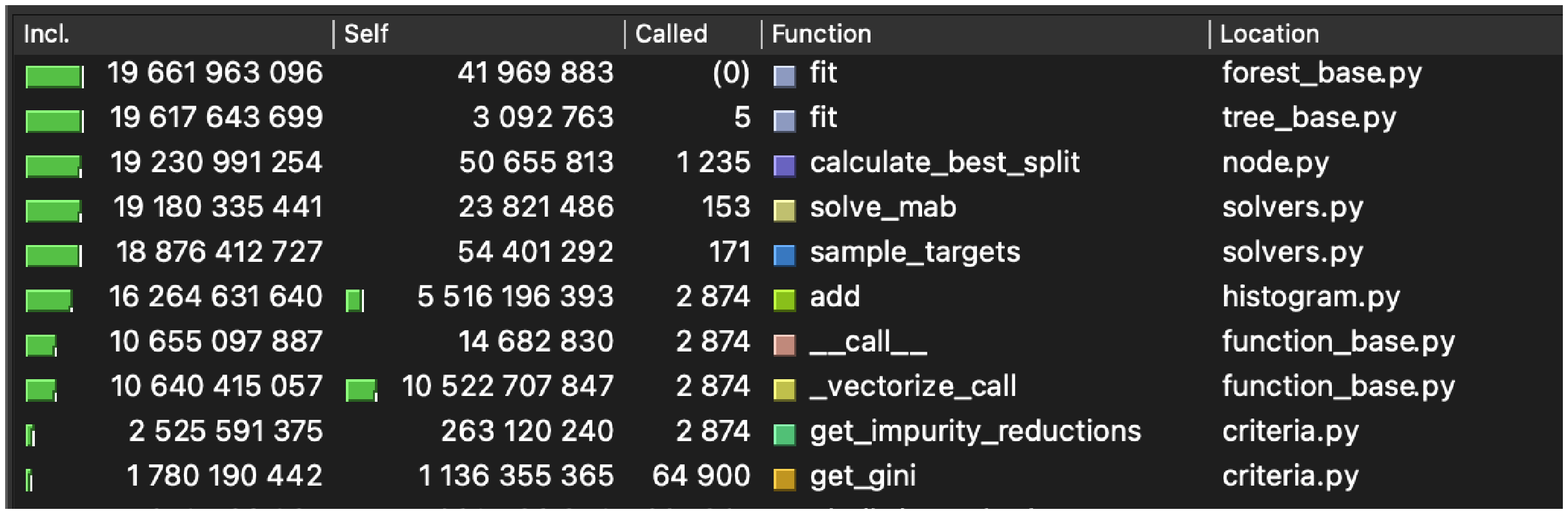}  
      \caption{}
    \end{subfigure}
    \begin{subfigure}{.9\textwidth}
      \centering
      \includegraphics[width=\linewidth]{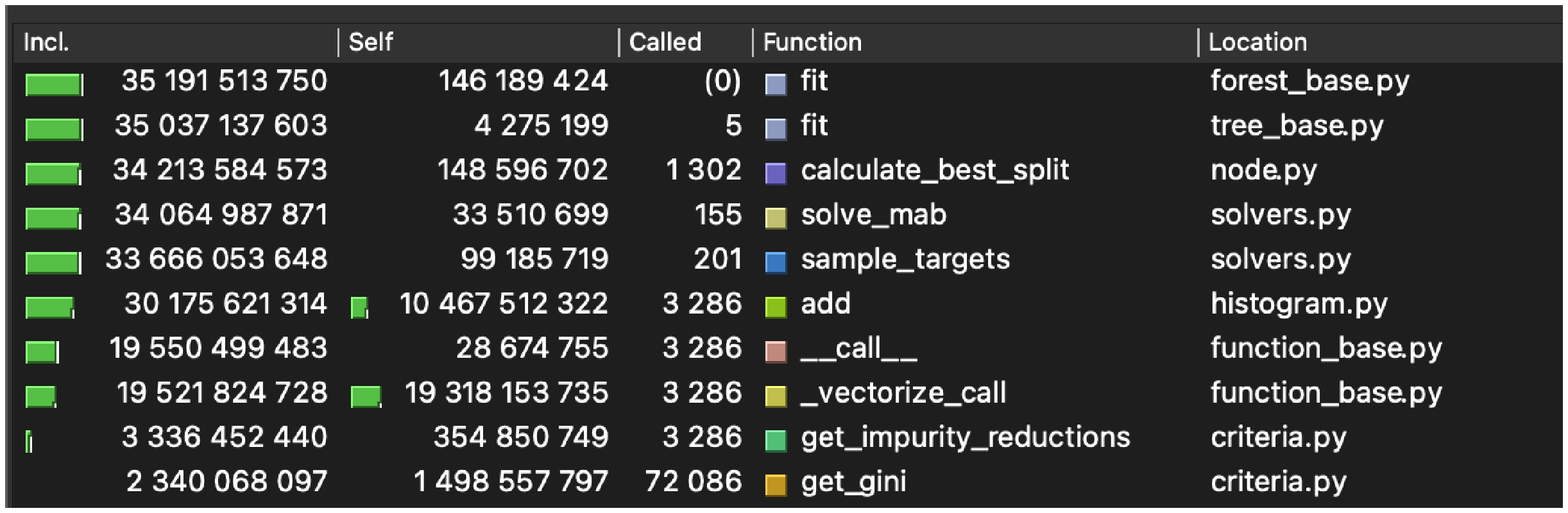}   
      \caption{}
    \end{subfigure}
    \caption{Profiles for the node-splitting algorithm using \algnamenospace, for 5,000 (top) and 10,000 (bottom) datapoint subsets of MNIST. The "Function" column is the name of the called function, the "Incl." column is the time spent in the function and any called subroutines, and the "Self" column is the time (in nanoseconds) spent in only the function and \textit{not} in any called sub-routines. All times are in nanoseconds. When increasing the dataset size, the time spent in the $\texttt{solve\_mab}$ function and any called subroutines only grows from approximately 20 seconds to approximately 35 seconds to identify the best feature-threshold split. This profile was generated with $\texttt{cProfile}$ and visualized with \texttt{pyprof2calltree} \cite{lanaro2019python}.}
    \label{fig:ourprofiles}
\end{figure}

Appendix Figure \ref{fig:theirprofiles} demonstrates the wall-clock time spent inside various functions when fitting a Random Forest classifier without \algname on two subsets of the MNIST dataset of sizes 5,000 and 10,000. Most of the time is spent inside the computation of the best feature-threshold split, which scales approximately as dataset size and motivates our focus on improving the performance of the split-identification subroutine. When using \algnamenospace, the time spent to identify the best feature-threshold split is reduced significantly (Appendix Figure \ref{fig:ourprofiles}).

\begin{figure}[h]
    \centering
    \includegraphics[scale=0.5]{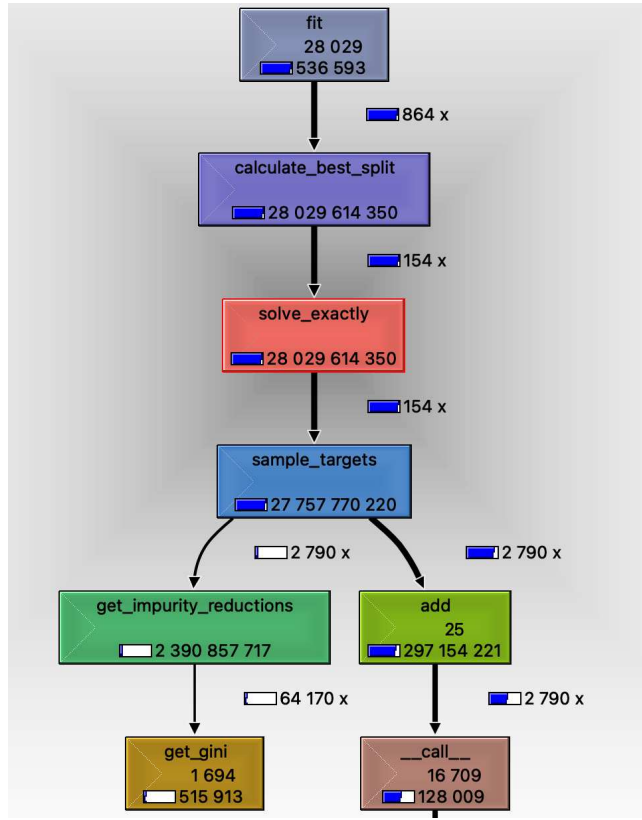}
    \caption{Example call graph of the $\texttt{fit}$ subroutine for the forest-based models in our re-implementation when the forest includes a single tree to be split only once. The $\texttt{fit}$ method of the forest calls the $\texttt{fit}$ method of its only tree, which calls $\texttt{calculate\_best\_split}$ method of the root node, which calls the respective solver ($\texttt{solve\_exactly}$ for the brute-force algorithm or $\texttt{solve\_exactly}$ for \algnamenospace), where the majority of wall-clock time is spent.}
\label{fig:callgraph}
\end{figure}

Appendix Figure \ref{fig:callgraph} also contains an example callgraph demonstrating callers and callees for the fitting procedure of a Random Forest, for easier interpretation of Appendix Figures \ref{fig:theirprofiles} and \ref{fig:ourprofiles}.

\clearpage
\section{Experiment Details}
\label{appendix:experimentdetails}

Here we provide full details for the experiments in Section \ref{sec:experiments}. All experiments were run on 2021 MacBook Pro running MacOS 12.5.1 (Monterey) with an Apple M1 Max processor, and 64 GB RAM.

\subsection{Datasets}

\paragraph{Classification Datasets: \label{subsec:class_data}} We use the MNIST \cite{lecun1998gradient}, APS Failure at Scania Trucks \cite{Gondek2016Predi-37212, Dua:2019}, and Forest Cover Type \cite{covtypedataset, Dua:2019} datasets. The MNIST dataset consists of 60,000 training and 10,000 test images of handwritten digits, where each black-and-white image is represented as a 784-dimensional vector and the task is to predict the digit represented by the image.
The APS Failure at Scania Trucks dataset consists of 60,000 datapoints with 171 features and the task is to predict component failure.
The Forest Covertype dataset consists of 581,012 datapoints with 54 feature and the task is to predict the type the forest cover type from cartographic variables.

\paragraph{Regression Datasets: \label{subsec:reg_data}} We use the Beijing Multi-Site Air-Quality \cite{Zhang2017CautionaryTO, Dua:2019} and the SGEMM GPU Kernel Performance \cite{GPU-1, GPU-2, Dua:2019} datasets. The Beijing Multi-Site Air-Quality dataset consists of 420,768 datapoints with 18 features and the task is to predict the level of air pollution. The SGEMM GPU Kernel Performance dataset consists of 241,600 datapoints and the task is to predict the running time of a matrix multiplication.

For all datasets except MNIST (which has predefined training and test datasets), all datasets were randomized into 9:1 train-test splits. All datasets are publicly available.

\subsection{Runtime Experiments}

For the runtime experiments presented in Tables \ref{table:classificationruntime}, all performances were measured from 5 random seeds. For all datasets, the maximum depth was set to 1 except for the MNIST dataset, in which the maximum depth was set to 5. The number of trees in each model was set to 5. All experiments used the Gini impurity criterion and the minimum impurity decrease required from performing a split was set to 0.005. For the Random Patches (RP) model, $\alpha_n$ was set to 0.7 and $\alpha_f$ was set to 0.85. 

For the regression runtime experiments presented in Table \ref{table:regressionruntime}, all performances were measured from 5 random seeds. For the Beijing Multi-Site Air-Quality Dataset, the maximum depth was set to 1 and for the SGEMM GPU Kernel Performance Dataset, the maximum number of leaf nodes was set to 5. The number of trees in each model was set to 5. All experiments used the MSE impurity criterion and the minimum impurity decrease required from performing a split was set to 0.005. For the Random Patches (RP) model, $\alpha_n$ was set to 0.7 and $\alpha_f$ was set to 0.85. 

\subsection{Budget Experiments}

For the classification budget experiments presented in Table \ref{table:classificationbudget}, all performances were measured from 5 random seeds. The budget for each model on the MNIST, APS Failure at Scania Trucks, and Forest Covertype datasets were set to 10,192,000, 784,000, and 9,408,000, respectively.  For the Random Patches (RP) model, $\alpha_n$ was set to 0.6 and $\alpha_f$ was set to 0.8. The maximum number of trees in any model was set to 100 and the maximum depth of each tree was set to 5.

For the regression budget experiments presented in Table \ref{table:regressionbudget}, all performances were measured from 5 random seeds. The budget for each model on the Beijing Multi-Site Air-Quality Dataset was set to 76,800,000 and the budget for each model on the SGEMM GPU Kernel Performance Dataset was set to 24,000,000. For the Random Patches (RP) model, $\alpha_n$ was set to 0.8 and $\alpha_f$ was set to 0.5. The maximum number of trees in any model was set to 100 and the maximum depth of each tree was set to 5.

\subsection{Stability Experiments}

Two metrics for calculating feature importance  are used in Table \ref{table:featureimportance}: out-of-bag Permutation Importance (OOB PI) and Mean Decrease in Impurity (MDI) \cite{nicodemus2011featurestability, piles2021piggenes}.
For a feature $f$, the OOB PI is calculated by measuring the difference between the trained model's out-of-bag error on the original data with its out-of-bag error on all the data with all out-of-bag datapoints' $f$ values shuffled.
The MDI for a feature $f$ is the average decrease in impurity of all nodes where $f$ is selected as the splitting criterion.

Once feature importances have been calculated, the top $k$ most important features for the model are selected and the stability of these $k$ features is measured via standard stability formulas \cite{nogueira2017stability}.

The results of the stability experiments are shown in Table \ref{table:featureimportance}.
The Random Classification dataset is generated via \texttt{scikit-learn}'s \texttt{datasets.make\_classification} function with $\texttt{n\_samples=10000, n\_features=60}$, and $\texttt{n\_informative=5}$.
The Random Regression dataset is generated by \texttt{scikit-learn}'s \texttt{datasets.make\_regression} with $\texttt{n\_samples=10000, n\_features=100}$, and $\texttt{n\_informative=5}$.

\clearpage
\section{Limitations}
\label{appendix:limitations}

\subsection{Theoretical Limitations}
\label{appendix:theoretical_limitations}

Crucial to the success of \algname are the assumptions described before and after Theorem \ref{thm:specific}. In particular, we assume that their is reasonable heterogeneity amongst the true impurity reductions of different feature-value splits. Such assumptions are common in the literature and have been validated on many real-world datasets \cite{bagaria2018medoids,zhang2019adaptive,baharav2019ultra,tiwari2020banditpam,bagaria2021bandit,baharav2022approximate}.

We also note that the assumptions that each CI scales as $\sqrt{\tfrac{\log 1/\delta}{n'}}$ may be violated when using certain impurity metrics. For example, the derivative of the entropy impurity criterion with respect to some $p_k$ approaches $\infty$ when $p_k \rightarrow 0$. In this case, we cannot apply the delta method from Appendix \ref{appendix:delta} to compute finite CIs that scale in the way we require. In such settings, it may be necessary to compute the CIs in other ways, e.g., following \cite{paninski2003entropyci} or \cite{basharin1959entropyci}.

We note that in the worst case, even when all assumptions are violated, MABSplit is never worse than the na\"ive algorithm in terms of sample complexity. In the worst case, it is a batched version of the na\"ive algorithm.

\subsection{Practical Limitations}
\label{appendix:practical_limitations}

We note that \algname may perform worse than na\"ive node-splitting on very small datasets, where the overhead of sampling the data in batches outweighs any potential benefits in sample complexity (see Appendix \ref{appendix:small_datasets} for further discussion).

In this work, we avoided a direct runtime comparison with \texttt{scikit-learn} because \texttt{scikit-learn} utilizes a number of low-level implementation optimizations that would make the comparison unfair. To provide a brief comparison to the popular \texttt{scikit-learn} implementation, however, we attempted to optimize our implementation using \texttt{numba} \cite{numba}, a package that translates Python code to optimized machine code. Our \texttt{numba}-optimized implementation is 4x faster than \texttt{scikit-learn}'s \texttt{DecisionTreeClassifier} and achieves comparable performance on the MNIST dataset; see Appendix Table \ref{table:sklearn_time_comparison}. 

\begin{table}[]
\resizebox{\textwidth}{!}{
\begin{tabular}{|ccc|}
\hline
\multicolumn{3}{|c|}{MNIST Dataset (Classification,  $N= 60,000$, maximum depth $= 8$)}                                                       \\
\multicolumn{1}{|c|}{Model}                                              & \multicolumn{1}{c|}{Wall-clock Training Time (s)} & Accuracy (\%)  \\ \hline
\multicolumn{1}{|c|}{ \texttt{scikit-learn} Decision Tree Classifier}     & \multicolumn{1}{c|}{34.665 ± 1.266}               & 91.061 ± 0.0   \\
\multicolumn{1}{|c|}{Histogrammed decision tree (Exact solver, ours)}    & \multicolumn{1}{c|}{86.514 ± 2.839}               & 90.923 ± 0.0   \\
\multicolumn{1}{|c|}{\textbf{Histogrammed decision tree (MABSplit solver, ours)}} & \multicolumn{1}{c|}{\textbf{8.538 ± 0.079} }               & \textbf{90.629 ± 0.234} \\ \hline
\end{tabular}
}
\vspace{0.2pt}
\caption{\label{table:sklearn_time_comparison}Comparison of accuracy and wall-clock training time of \texttt{scikit-learn}'s Decision Tree Classifier with our implementation on the MNIST digit classification task. Our implementation of the histogrammed decision tree is slower than \texttt{scikit-learn}'s, but our optimized implementation is about 4x faster than \texttt{scikit-learn}'s. The slight performance degradation is likely due to discretization of the data during histogramming; this effect is also seen when histogramming the data and using the exact solver (i.e., when not using \algnamenospace).  A more heavily optimized version of our histogrammed decision tree when using \algname would likely result in even lower training times. Performance was measured over 5 random seeds.}
\end{table}

In order for practitioners to take full advantage of \algnamenospace, however, it may be necessary to implement \algname within the \texttt{scikit-learn} library.
In doing so, it may be possible that \algname makes it difficult or impossible to use existing optimizations in the \texttt{scikit-learn} library. An example of this is vectorization: because the na\"ive node-splitting algorithm queries the data in a predictable way, each datapoint can be queried more quickly than in \algnamenospace. Despite \algnamenospace's advantages in sample complexity, the disadvantages of being unable to use implementation optimizations like vectorization may outweigh \algnamenospace's benefits. 
Many of these risks may be ameliorated by addressed \algname into existing RF implementations such as the one in \texttt{scikit-learn}. We anticipate that many optimizations will still apply: for example pre-fetching data to have it in caches close to the CPU, manual loop unrolling, etc. 
We leave an optimization implementation of \algname inside the \texttt{scikit-learn} library to future work.

\clearpage
\section{Comparison on Small Datasets}
\label{appendix:small_datasets}

\begin{center}
\begin{figure}[h!]
    \begin{subfigure}{.95\textwidth}
        \centering
        \includegraphics[width=\linewidth]{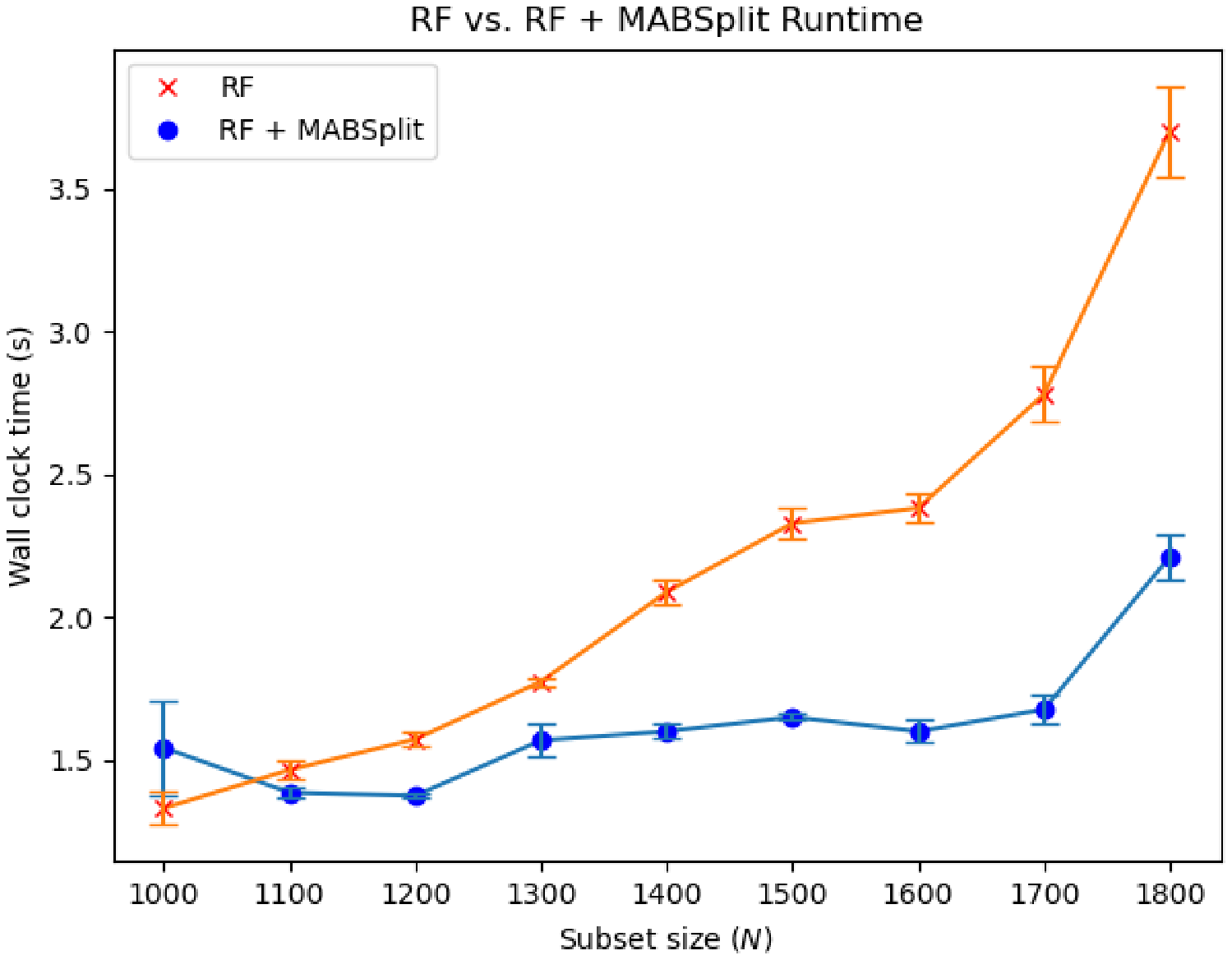} 
        \caption{}
    \end{subfigure}
    \begin{subfigure}{.95\textwidth}
      \centering
      \includegraphics[width=\linewidth]{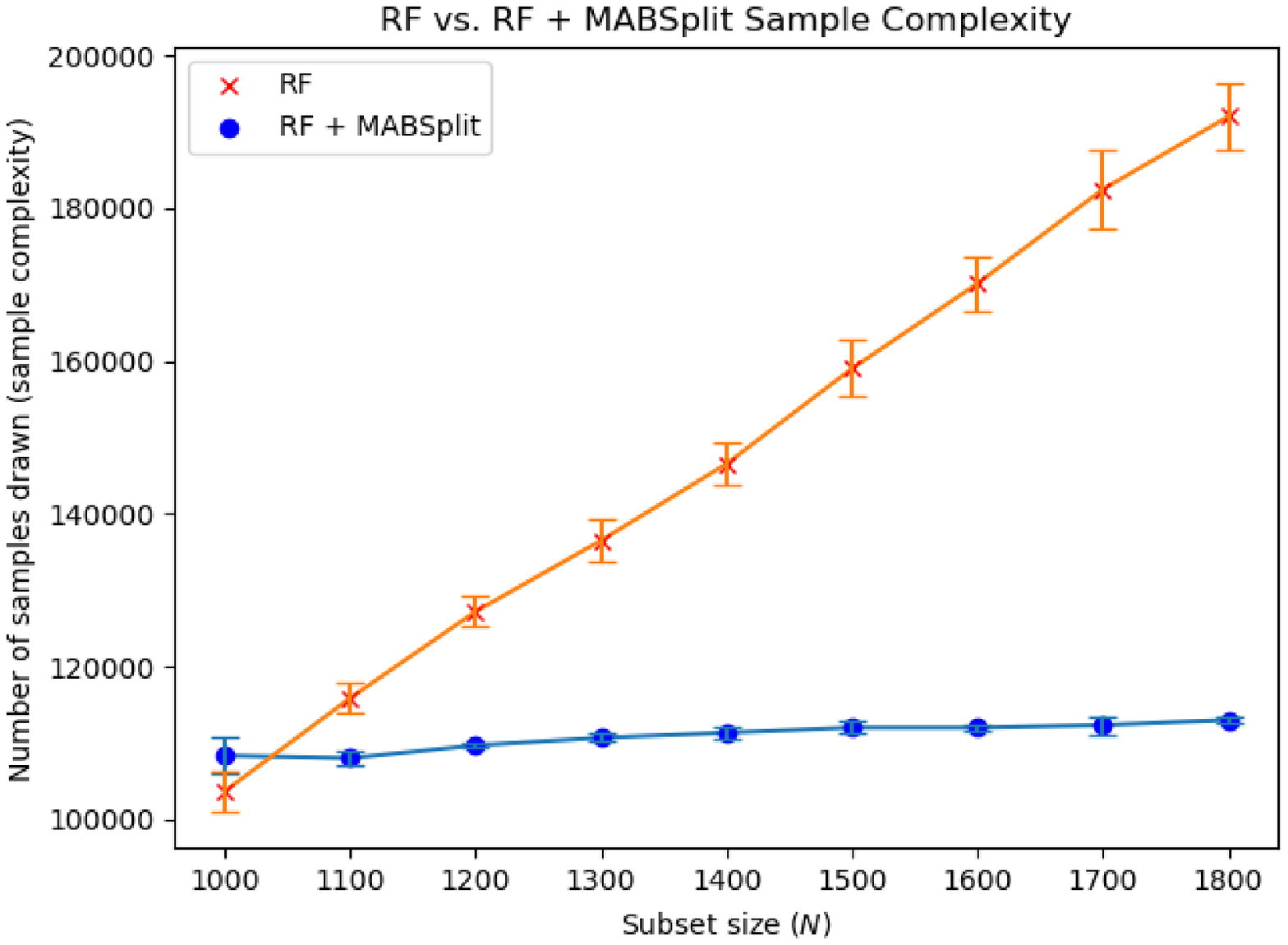}   
      \caption{}
    \end{subfigure}
    \caption{(a) Wall-clock training times and (b) sample complexities of a random forest model with and without MABSPlit, for various subset sizes of MNIST. For dataset sizes below approximately $1000$, the exact random forest model performs better in terms of sample complexity and wall-clock time. Above $1100$ datapoints, the MABSplit version demonstrates better sample complexity and wall-clock time. Error bars were computed over 3 random seeds. Test performances were not different at a statistically significant level.}
    \label{fig:small_exps_runtime}
\end{figure}
\end{center}

In this section, we investigate the performance of \algname on small datasets. Appendix Figure \ref{fig:small_exps_runtime} demonstrates the performance of \algnamenospace, both in wall-clock training time and sample complexity, for various subset sizes of MNIST. Our results that RF+MABSplit outperforms the standard RF algorithm, in both sample complexity and wall-clock time, when the dataset size exceeds approximately $1100$ datapoints. 

However, we also note that the main use case for MABSplit is when the data size is large and it is computationally challenging to run standard forest-based algorithms. Indeed, the use of big data in many applications that necessitate sampling was the primary motivation for our work \cite{selfGFN, selfBIG, selfNLA, selfEye, selfQuantum, selfGoogle}.

\clearpage
\section{Description of Other Node-Splitting Algorithms}
\label{app:node_splitting_algs}

For completeness, we provide a brief description of various baseline models' node-splitting algorithms here to enable easier comparison with \algnamenospace.

Consider a node with $n$ datapoints each with $m$ features, and $T$ possible thresholds at which to split each feature. We discuss the classification setting for simplicity, though the same arguments apply to regression.

A very na\"ive approach would be to iterate over all $mT$ feature-value splits, and compute the probabilities $p_{L, k}$ and $p_{R,k}$ from all $n$ datapoints. This results in complexity $O(mTn)$, which is $O(mn^2)$ when $T = n$ (for example, $T = n$ in the un-histogrammed setting).

Instead, the usual RF algorithm sorts all $n$ datapoints in $O(n\text{log}n)$ time for each of the $m$ features, resulting in total computational cost $O(mn\text{log}n)$. Then the algorithm scans linearly from lowest value to highest value for each feature and update the parameters $p_{L, k}$ and $p_{R,k}$ via simple counting to find the best impurity reduction for each of the $T$ potential splits. The complexity of this step is $O(mT + mn)$, where the ``$+mn$'' comes from the allocations of each data point to the left or right node during the scan (each data point is re-allocated only once per feature). Thus the total complexity of this approach is $O(mn\text{log}n + mT + mn) = O(mn\text{log}n + mT)$. This is $O(mn\text{log}n)$ when $T = n$.

The binned (a.k.a. histogrammed) method does not require the per-feature sort and avoids the $O(mn\text{log}n)$ computation when $T < n$,. Instead, each of the $n$ points must be inserted into the correct bin (which can be done in $O(1)$ time for each datapoint if the bins are equally spaced) for each of the $m$ features, incurring total computational cost $O(mn)$. Then, the same linear scanning approach as in the ``standard'' algorithm is performed with complexity $O(mT + mn)$. The total complexity of this approach is $O(mn + mT + mn) = O(m(n+T))$. This is $O(mn)$ when $T = n$.

In general, we do not assume $T = n$, i.e., that every feature value is a potential split point, unless otherwise specified. In our paper, the ``standard'' approach refers to the \underline{\textbf{un}}binned approach which requires an $O(mn\text{log}n)$ sort and ``linear'' refers to the binned approach that is $O(m(n+T))$, which is $O(mn)$ when $ T = O(n)$.

Crucially, when $T = o(N)$ (as is often the case in practice, e.g., for a constant number of bins) and the necessary gap assumptions are satisfied, MABSplit scales as $O(mT\text{log}n)$. In many cases, this is much better than $O(m(n+T))$, e.g., for large datasets, because the dependence on $n$ is reduced from linear to logarithmic. More concretely, treating $T$ as a constant and ignoring the dependence on $m$, we reduce the complexity of the binned algorithm from $O(n)$, what we refer to as ``linear,'' to $O(\text{log}n)$.

\end{document}